\documentclass[letterpaper]{article} % DO NOT CHANGE THIS
\usepackage{aaai23}
\usepackage{times}  % DO NOT CHANGE THIS
\usepackage{helvet}  % DO NOT CHANGE THIS
\usepackage{courier}  % DO NOT CHANGE THIS
\usepackage[hyphens]{url}  % DO NOT CHANGE THIS
\usepackage{graphicx} % DO NOT CHANGE THIS
\usepackage{natbib}  % DO NOT CHANGE THIS AND DO NOT ADD ANY OPTIONS TO IT
\usepackage{caption} % DO NOT CHANGE THIS AND DO NOT ADD ANY OPTIONS TO IT
\usepackage{algorithm}
\usepackage{algorithmic}
\usepackage{soul}

\usepackage{amsmath}
\usepackage{amsthm}
\usepackage{amssymb}
\usepackage{enumitem}
\usepackage{color}
\usepackage{subfigure}
\usepackage{multirow}
\usepackage{booktabs}       % professional-quality tables
\usepackage{xcolor}

\newtheorem{definition}{Definition}[section]
\newtheorem{theorem}{Theorem}[section]

\newtheorem{lemma}{Lemma}[section]
\newtheorem{problem}{Problem}

\usepackage{newfloat}
\usepackage{listings}
\DeclareCaptionStyle{ruled}{labelfont=normalfont,labelsep=colon,strut=off} % DO NOT CHANGE THIS
\floatstyle{ruled}
\newfloat{listing}{tb}{lst}{}
\floatname{listing}{Listing}
\usepackage{hyperref}

\title{Stable Learning via Sparse Variable Independence}
\author{
    Han Yu\textsuperscript{\rm 1}, 
    Peng Cui\textsuperscript{\rm 1}\thanks{Corresponding author}, 
    Yue He\textsuperscript{\rm 1}, 
    Zheyan Shen\textsuperscript{\rm 1}, 
    Yong Lin\textsuperscript{\rm 2}, 
    Renzhe Xu\textsuperscript{\rm 1}, 
    Xingxuan Zhang\textsuperscript{\rm 1}
}
\affiliations{
    \textsuperscript{\rm 1}Tsinghua University, 
    \textsuperscript{\rm 2}Hong Kong University of Science and Technology
    yuh21@mails.tsinghua.edu.cn, cuip@tsinghua.edu.cn, heyue18@mails.tsinghua.edu.cn\\ shenzy13@qq.com, ylindf@connect.ust.hk, xrz199721@gmail.com, xingxuanzhang@hotmail.com
}

\usepackage{bibentry}
\begin{document}

\maketitle

\begin{abstract}
The problem of covariate-shift generalization has attracted intensive research attention. 
Previous stable learning algorithms employ sample reweighting schemes to decorrelate the covariates when there is no explicit domain information about training data.
However, with finite samples, it is difficult to achieve the desirable weights that ensure perfect independence to get rid of the unstable variables. 
Besides, decorrelating within stable variables may bring about high variance of learned models because of the over-reduced effective sample size. 
A tremendous sample size is required for these algorithms to work. 
In this paper, with theoretical justification, we propose SVI (Sparse Variable Independence) for the covariate-shift generalization problem. We introduce sparsity constraint to compensate for the imperfectness of sample reweighting under the finite-sample setting in previous methods. 
Furthermore, we organically combine independence-based sample reweighting and sparsity-based variable selection in an iterative way to avoid decorrelating within stable variables, increasing the effective sample size to alleviate variance inflation. 
Experiments on both synthetic and real-world datasets demonstrate the improvement of covariate-shift generalization performance brought by SVI. 
\end{abstract}

\section{Introduction}

Most of the current machine learning techniques rely on the IID assumption, that the test and training data are independent and identically distributed, which is too strong to stand in wild environments \cite{koh2021wilds}. 
Test distribution often differs from training distribution especially if there is data selection bias when collecting the training data \cite{Heckman1979SampleSB,Young2009GeneOA}. 
Covariate shift is a common type of distribution shift \cite{Ioffe2015BatchNA,Tripuraneni2021OverparameterizationIR}, which assumes that the marginal distribution of covariates (i.e. $P(\boldsymbol{X})$) may shift between training and test data while the generation mechanism of the outcome variable (i.e. $P(Y|\boldsymbol{X})$) remains unchanged \cite{shen2021towards}. 
To address this problem, there are some strands of works \cite{Santurkar2018HowDB,Wilson2020ASO,Wang2021GeneralizingTU}. 
When some information about the test distribution is known apriori \cite{Peng2019MomentMF,Yang2020FDAFD}, domain adaptation methods are proposed based on feature space transformation or distribution matching \cite{ben2010theory, weiss2016survey,Tzeng2017AdversarialDD,Ganin2015UnsupervisedDA,Saito2018MaximumCD}. 
If there exists explicit heterogeneity in the training data, e.g. it is composed of multiple subpopulations corresponding to different source domains \cite{Blanchard2021DomainGB,Gideon2021ImprovingCS}, domain generalization methods are proposed to learn a domain-agnostic model or invariant representation \cite{muandet2013domain, li2017deeper,ganin2016domain,li2018domain,sun2016deep, he2021towards, Zhang2022NICOTB, zhang2022towards}. 
In many real applications, however, neither the knowledge about test data nor explicit domain information in training data is available.

Recently, stable learning algorithms \cite{shen2018causally, shen2020stable2, shen2020stable, kuang2018stable, kuang2020stable, zhang2021deep, liu2021heterogeneous, liu2021kernelized} are proposed to address a more realistic and challenging setting, where the training data consists of latent heterogeneity (without explicit domain information), and the goal is to achieve a model with good generalization ability under agnostic covariate shift. 
They make a structural assumption of covariates by splitting them into $\boldsymbol{S}$ (i.e. stable variables) and $\boldsymbol{V}$ (i.e. unstable variables), and suppose $P(Y|\boldsymbol{S})$ remains unchanged while $P(Y|\boldsymbol{V})$ may change under covariate shift.
They aim to learn a group of sample weights to remove the correlations among covariates in observational data, and then optimize in the weighted distribution to capture stable variables.
It is theoretically proved that, under the scenario of infinite samples, these models only utilize the stable variables for prediction (i.e. the coefficients on unstable variables will be perfectly zero) if the learned sample weights can strictly ensure the mutual independence among all covariates \cite{xu2021stable}. 
However, with finite samples, it is almost impossible to learn weights that satisfy complete independence. 
As a result, the predictor cannot always get rid of the unstable variables (i.e. the unstable variables may have significantly non-zero coefficients). 
In addition, \cite{shen2020stable2} pointed out that it is unnecessary to remove the inner correlations of stable variables. 
The correlations inside stable variables could be very strong, thus decorrelating them could sharply decrease the effective sample size, even leading to the variance inflation of learned models. Taking these two factors together, the requirement of a tremendous effective sample size severely restricts the range of applications for these algorithms.

In this paper, we propose a novel algorithm named Sparse Variable Independence (SVI) to help alleviate the rigorous requirement of sample size.
We integrate the sparsity constraint for variable selection and the sample reweighting process for variable independence into a linear/nonlinear predictive model. 
We theoretically prove that even when the data generation is nonlinear, the stable variables can surely be selected with a sparsity constraint like $\ell_1$ penalty, if the correlations between stable variables and unstable variables are weak to some extent. 
Thus we do not require complete independence among covariates. 
To further reduce the requirement on sample size, we design an iterative procedure between sparse variable selection and sample reweighting to prevent attempts to decorrelate within stable variables. 
The experiments on both synthetic and real-world datasets
clearly demonstrate the improvement of covariate-shift generalization performance brought by SVI.  

The main contributions in this paper are listed below:
\begin{itemize}
    \item  We introduce the sparsity constraint to attain a more pragmatic independence-based sample reweighting algorithm which improves covariate-shift generalization ability with finite training samples. We theoretically prove the benefit of this. 
    \item We design an iterative procedure to avoid decorrelating within stable variables, mitigating the problem of over-reduced effective sample size. 
    \item We conduct extensive experiments on various synthetic and real-world datasets to verify the advantages of our proposed method.
\end{itemize}

\section{Problem Definition}
\emph{Notations}: Generally in this paper, a bold-type letter represents a matrix or a vector, while a normal letter represents a scalar. 
Unless otherwise stated, $\boldsymbol{X}\in \mathbb{R}^p$ denotes the covariates with dimension of $p$, $X_d$ denotes the d$^{th}$ variable in $\boldsymbol{X}$, and $Y\in \mathbb{R}$ denotes the outcome variable. 
Training data is drawn from the distribution $P^{tr}(\boldsymbol{X}, Y)$ while test data is drawn from the unknown distribution $P^{te}(\boldsymbol{X}, Y)$. 
Let $\mathcal{X}, \mathcal{X}_d, \mathcal{Y}$ denote the support of $\boldsymbol{X}, X_d, Y$ respectively. 

$\boldsymbol{S}\subseteq \boldsymbol{X}$ implies $\boldsymbol{S}$ is a subset variables of $\boldsymbol{X}$ with dimension $p_s$. We use $\boldsymbol{A}\perp \boldsymbol{B}$ to denote that $\boldsymbol{A}$ and $\boldsymbol{B}$ are statistically independent of each other. 
We use $\mathbb{E}_{Q(\cdot)}[\cdot]$ to denote expectation and $\mathbb{E}_{Q(\cdot)}[\cdot|\cdot]$ to denote conditional expectation under distribution $Q$, which can be chosen as $P^{tr}$, $P^{te}$ or any other proper distributions.

% \vspace{10pt}

In this work, we focus on the problem of covariate shift. 
% which assumes that the distribution shifts occur on the marginal distribution of covariates $\boldsymbol{X}$ while the conditional distribution of the outcome variable $Y$ given $\boldsymbol{X}$ remains unchanged. 
It is a typical and most common kind of distribution shift considered in OOD literature.

% For a general machine learning problem, given the features $\boldsymbol{X}$ and the outcome $Y$, we aim to learn a model to predict $Y$ from $\boldsymbol{X}$. 
% We focus on OOD problems with covariate shift, which assumes that distribution shifts occur on the marginal distribution of $\boldsymbol{X}$ with the conditional distribution of $Y$ given $\boldsymbol{X}$ remaining unchanged. 
% It is a typical and common kind of distribution shift among all possible ones \cite{shen2021towards}.

% \begin{assumption} [Covariate shift] \label{assum:cs}

% Suppose test distribution $P^{te}$ differs from training distribution $P^{tr}$ in the shift of covariates' distribution only, \textit{i.e.}, 
%     \begin{equation}
%         P^{te}(\boldsymbol{X}, Y)=P^{te}(\boldsymbol{X})P^{tr}(Y|\boldsymbol{X})
%     \end{equation}
%     Moreover, $P^{te}$ shares the same support with $P^{tr}$.
% \end{assumption}

% \begin{assumption} [Strictly positive density] \label{assum:positive}
%     $\forall x_1 \in \mathcal{X}_1, x_2 \in \mathcal{X}_2, \dots, x_p \in \mathcal{X}_p$, $P^{tr}(X_1 = x_1, X_2 = x_2, \dots, X_p = x_p) > 0$.
% \end{assumption}

% Further, we make the Assumption \ref{assum:positive} which requires strictly positive probability density on the support of $\boldsymbol{X}$, ensuring the feasibility of generalization from training
% distribution to agnostic test distribution.

\begin{problem}[Covariate-Shift Generalization]
    Given the samples $\{(\boldsymbol{X}_i, Y_i)\}_{i=1}^{N}$ drawn from training distribution $P^{tr}$, the goal of covariate-shift generalization is to learn a prediction model so that it performs stably on predicting $Y$ in agnostic test distribution where $P^{te}(\boldsymbol{X}, Y)=P^{te}(\boldsymbol{X})P^{tr}(Y|\boldsymbol{X})$.
\end{problem}

To address the covariate-shift generalization problem, we define minimal stable variable set \cite{xu2021stable}.

\begin{definition} [Minimal stable variable set] 
    A minimal stable variable set of predicting $Y$ under training distribution $P^{tr}$ is any subset $\boldsymbol{S}$ of $\boldsymbol{X}$ satisfying the following equation, and none of its proper subsets satisfies it.
    
    \begin{equation}
    \small
        \mathbb{E}_{P^{tr}}[Y | \boldsymbol{S}] = \mathbb{E}_{P^{tr}}[Y | \boldsymbol{X}]. \label{eq:stable-set}
    \end{equation}
    
\end{definition}

Under strictly positive density assumption, the minimal stable variable set $\boldsymbol{S}$ is unique. In the setting of covariate shift where $P^{tr}(\boldsymbol{X})\neq P^{te}(\boldsymbol{X})$, relationships between $\boldsymbol{S}$ and $\boldsymbol{X}\backslash\boldsymbol{S}$ can arbitrarily change, resulting in the unstable correlations between $Y$ and $\boldsymbol{X}\backslash\boldsymbol{S}$. Demonstrably, 
according to \cite{xu2021stable}, $\boldsymbol{S}$ is a minimal and optimal predictor of $Y$ under test distribution $P^{te}$ if and only if it is a minimal stable variable set under $P^{tr}$. Hence, in this paper, we intend to capture the minimal stable variable set $\boldsymbol{S}$ for stable prediction under covariate shift.
Without ambiguity, we refer to $\boldsymbol{S}$ as stable variables and $\boldsymbol{V}=\boldsymbol{X}\backslash\boldsymbol{S}$ as unstable variables in the rest of this paper.

% We consider the classic setting  stable learning, assuming the following data generation process:
% \begin{equation}
%     \boldsymbol{X}=(\boldsymbol{S}, \boldsymbol{V}), \quad Y=f(\boldsymbol{S})+\epsilon, \quad \epsilon \perp \boldsymbol{X}
% \end{equation}

% It is easy to find that here MinStable$(Y)$ is the direct causal variables $\boldsymbol{S}$, while relationship between $\boldsymbol{S}$ and $\boldsymbol{V}$ can arbitrarily change. 
% Therefore, $\boldsymbol{S}$ is exactly the variable set we desire for. 

\section{Method}

\subsection{Independence-based sample reweighting}
First, we define the weighting function and the target distribution that we want the training distribution to be reweighted to.

\begin{definition} [Weighting function] \label{def:weighting}
    Let $\mathcal{W}$ be the set of weighting functions that satisfy
    \begin{equation}
    \small
        \mathcal{W} = \left\{w: \mathcal{X} \rightarrow \mathbb{R}^{+} \mid \mathbb{E}_{P^{tr}(\boldsymbol{X})}[w(\boldsymbol{X})] = 1 \right\}.
    \end{equation}
    Then $\forall w \in \mathcal{W}$, the corresponding weighted distribution is $\tilde{P}_w(\boldsymbol{X}, Y) = w(\boldsymbol{X})P^{tr}(\boldsymbol{X}, Y)$. $\tilde{P}_w$ is well defined with the same support of $P^{tr}$.
\end{definition}

Since we expect that variables are decorrelated in the weighted distribution, we denote $\mathcal{W}_{\perp}$ as the subset of $\mathcal{W}$ in which $\boldsymbol{X}$ are mutually independent in the weighted distribution $\tilde{P}_w$.

Under the infinite-sample setting, it is proved that if conducting weighted least squares using the weighting function in $\mathcal{W}_{\perp}$, almost surely there will be non-zero coefficients only on stable variables, no matter whether the data generation function is linear or nonlinear \cite{xu2021stable}. 
However, the condition for this to hold is too strict and ideal. 
Under the finite-sample setting, we can hardly learn the sample weights corresponding to the weighting function in $\mathcal{W}_\perp$. 

We now take a look at two specific techniques of sample reweighting, which will be incorporated into our algorithms.

\paragraph{DWR} \cite{kuang2020stable} aims to remove linear correlations between each two variables, \textit{i.e.},
\begin{equation} \label{eq:DWR}
\small
    \hat{w}(\boldsymbol{X}) = \arg \min_{w(\boldsymbol{X})} \sum_{1\le i,j \le p, i\ne j}\left(Cov(X_i, X_j; w)\right)^2,
\end{equation}
where $Cov(X_i, X_j; w)$ represents the covariance of $X_i$ and $X_j$ after being reweighted by $\tilde{P}_{w}$. 
DWR is well fitted for the case where the data generation process is dominated by a linear function, since it focuses on linear decorrelation only. 

\paragraph{SRDO} \cite{shen2020stable} conducts sample reweighting by density ratio estimation. 
It simulates the target distribution $\tilde{P}$ by means of random resampling on each covariate so that $\tilde{P}(X_1, X_2, \dots, X_p) = \prod_{i=1}^p P^{tr}(X_i)$.
Then the weighting function can be estimated by
\begin{equation} \label{eq:SRDO}
\small
    \hat{w}(\boldsymbol{X}) = \frac{\tilde{P}(\boldsymbol{X})}{P^{tr}(\boldsymbol{X})} =  \frac{P^{tr}(X_1)P^{tr}(X_2)\dots P^{tr}(X_p)}{P^{tr}(X_1, X_2, \dots, X_p)}.
\end{equation}
To estimate such density ratio, SRDO learns an MLP classifier to differentiate whether a sample belongs to the original distribution $P^{tr}$ or the mutually independent target distribution $\tilde{P}$. 
Unlike DWR, this method can not only decrease linear correlations among covariates, but it can weaken nonlinear dependence among them. 

Under the finite-sample setting, for DWR, if the scale of sample size is not significantly greater than that of covariates dimension, it is difficult for Equation \ref{eq:DWR} to be optimized close to zero. 
For SRDO, $\tilde{P}$ is generated by a rough process of resampling, further resulting in the inaccurate estimation of density ratio. 
In addition, both methods suffer from the over-reduced effective sample size when strong correlations exist inside stable variables, since they decorrelate variables globally.
Therefore, they both require an enormous sample size to work. 

\subsection{Sample Reweighting with Sparsity Constraint}

\paragraph{Motivation and general idea}

By introducing sparsity constraint, we can loose the demand for the perfectness of independence achieved by sample reweighting under the finite-sample setting, reducing the requirement on sample size. 
Inspired by \cite{zhao2006model}, we theoretically prove the benefit of this. 

\begin{theorem}
 
 \label{prop:lasso}

Set $\boldsymbol{X^{(n)}}=[\boldsymbol{S}^{(n)}, \boldsymbol{V}^{(n)}], \boldsymbol{\beta}=\begin{pmatrix}\boldsymbol{\beta}_s\\\boldsymbol{\beta}_v\end{pmatrix}$. 
For data generated by: $\boldsymbol{Y}^{(n)}=\boldsymbol{X}^{(n)}\boldsymbol{\beta}+ g^{(n)}(\boldsymbol{S})+\boldsymbol{\epsilon}^{(n)}= \boldsymbol{S}^{(n)} \boldsymbol{\beta}_s + \boldsymbol{V}^{(n)} \boldsymbol{\beta}_v + g^{(n)}(\boldsymbol{S}) + \boldsymbol{\epsilon}^{(n)}$, where $\boldsymbol{\epsilon}^{(n)}$ is a vector of i.i.d. random variables with mean $0$ and variance $\sigma^2$, $\boldsymbol{Y}^{(n)}$ is $n\times 1$ outcome, $\boldsymbol{S}^{(n)}$ and $\boldsymbol{V}^{(n)}$ are $n\times p_s$ and $n\times p_v$ data matrix respectively, $g^{(n)}(\boldsymbol{S})$ is $n\times 1$ nonlinear term, which is linearly uncorrelated with the linear part. 
$\boldsymbol{\beta}_s$ is regression coefficients for $p_s$ stable variables whose elements are all non-zero, while $\boldsymbol{\beta}_v=\boldsymbol{0}$.

Assume $\boldsymbol{S}$ and $\boldsymbol{V}$ are normalized to zero mean with finite second-order moment, and both the covariates and the nonlinear term are bounded, i.e. almost surely $||\boldsymbol{X}||_2\leq B, g(\boldsymbol{S})\leq \delta$. If there exists a positive constant vector $\boldsymbol{\eta}$ such that:
\begin{equation}
\label{eq:cov}
\small
|{\rm Cov}^{(n)}(\boldsymbol{V}, \boldsymbol{S}){\rm Cov}^{(n)}(\boldsymbol{S})^{-1}{\rm  sign}(\boldsymbol{\beta}_s)| \leq \boldsymbol{1} - \boldsymbol{\eta}    
\end{equation}
where ${\rm Cov}^{(n)}$ represents sample covariance, ${\rm sign}$ represents element-wise sign function. 

Then there exists a positive constant $M$ which can be written as $h(\delta, B, {\rm Cov}(\boldsymbol{X}))$ unrelated to $n$ for the following inequality to stand: $\forall \lambda_n$ satisfying $\lambda_n=o(n)$ and $\lambda_n=\omega\left(n^{\frac{1+c}{2}}\right)$ with $0\leq c \le 1$:
\begin{equation}
\small
    P(\hat{\boldsymbol{\beta}}(\lambda_n)=_s \boldsymbol{\beta})\geq 1-o\left(n^{-\frac{c}{2}}e^{-\frac{\min_{i=1}^{p_v}\{\eta_i\}^2}{8M^2}n^c}\right)
\end{equation}
where ``$=_s$" means equal after applying sign function, $\hat{\boldsymbol{\beta}}(\lambda_n)$ is the optimal solution of Lasso with regularizer coefficient $\lambda_n$. 
\end{theorem} 

The term ${\rm Cov}^{(n)}(\boldsymbol{V}, \boldsymbol{S})$ measures the dependency between stable variables and unstable variables. 
Thus Theorem \ref{prop:lasso} implies that if correlations between $\boldsymbol{S}$ and $\boldsymbol{V}$ are weakened to some extent, the probability of perfectly selecting stable variables $\boldsymbol{S}$ approaches 1 at an exponential rate. 
This confirms that adding sparsity constraint to independence-based sample reweighting may loose the requirement for independence among covariates, even when the data generation is nonlinear. 

\begin{figure}[t]
  \centering
  \includegraphics[width=\linewidth]{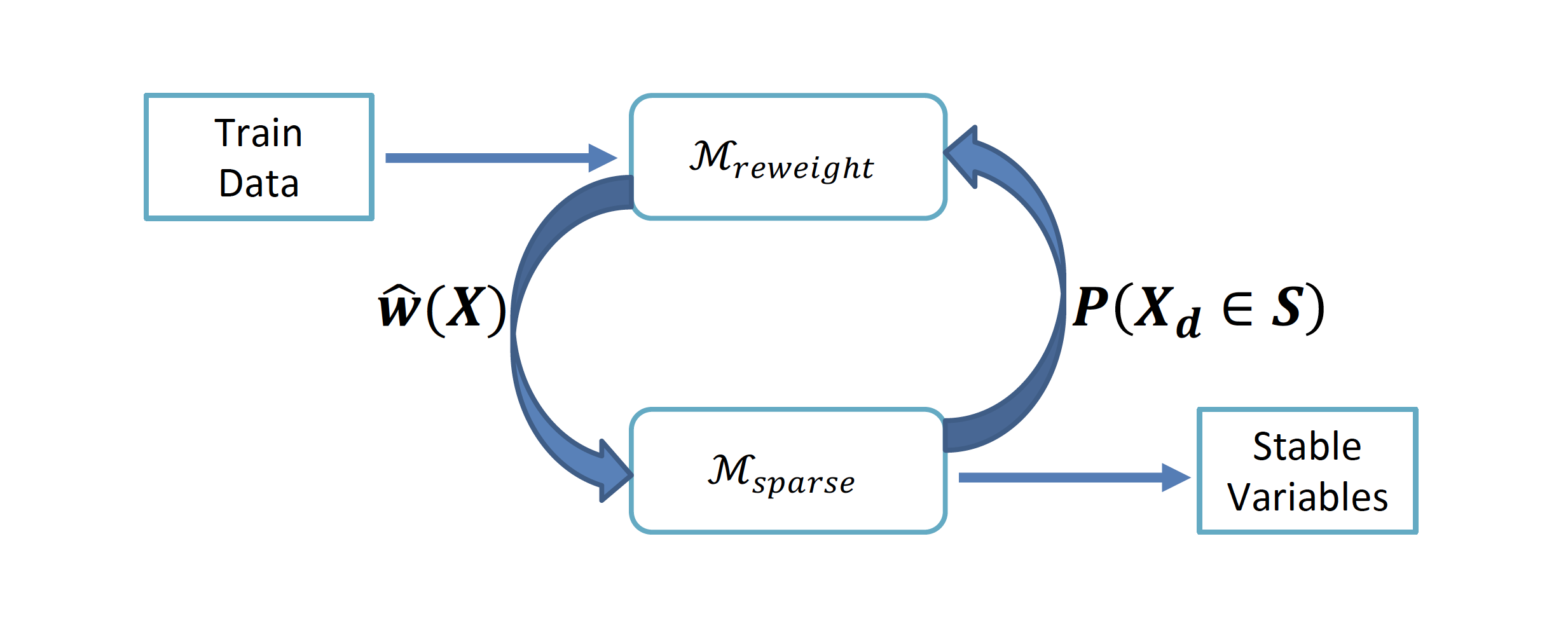}
  \caption{Diagram of SVI.}
%   \caption{Diagram of SRSC. The frontend module of sample reweighting for decorrelation and the backend module of sparse learning to capture stable variables can promote each other in an iterative loop. Better learned  $\hat{w}(\boldsymbol{X})$ can benefit the soft selection of stable variables, i.e., $P(\boldsymbol{X_d}\in \boldsymbol{S})$. Meanwhile, if the stable variables are selected more precisely, correlation among stable variables can be preserved, thus mitigating the problem of over-reduced effective sample size. }
  \label{fig:iter}
\end{figure}

Motivated by Theorem \ref{prop:lasso}, we propose the novel Sparse Variable Independence (SVI) method. The algorithm mainly consists of two modules: the frontend for sample reweighting, and the backend for sparse learning under reweighted distribution. We then combine the two modules in an iterative way as shown in Figure \ref{fig:iter}. Details are provided as follows.

\paragraph{Frontend implementation} We employ different techniques under different settings.  For the case where data generation is dominated by a linear function, we employ DWR, namely Equation \ref{eq:DWR} as loss function to be optimized using gradient descent. For the case where data generation is totally governed by nonlinear function, we employ SRDO, namely Equation \ref{eq:SRDO} to conduct density ratio estimation for reweighting.

\paragraph{Backend implementation}
As for the backend sparse learning module, in order to take the nonlinear setting into account, we implement the sparsity constraint following \cite{yamada2020feature} instead of Lasso. 
Typical for variable selection, we begin with an $\ell_0$ constraint which is equivalent to multiplying the covariates $\boldsymbol{X}$ with a hard mask $\boldsymbol{M}=[m_1, m_2, ...,m_p]^T $ whose elements are either one or zero. 
We approximate the elements in $\boldsymbol{M}$ using clipped Gaussian random variables parametrized by $\boldsymbol{\mu}=[\mu_1, \mu_2, ..., \mu_p]^T$:
\begin{equation}
\small
    m_d=\max\{0, \min\{1, \mu_d + \epsilon_d\}\}
\end{equation}
where $\epsilon_d$ is drawn from zero-mean Gaussian distribution $\mathcal{N}(0, \sigma^2)$.

The standard $\ell_0$ constraint for a general function $f$ parametrized by $\boldsymbol{\theta}$ can be written as:
\begin{equation} \label{eq:l0}
\small
    \mathcal{L}(\boldsymbol{\theta}, \boldsymbol{\mu})=\mathbb{E}_{P(\boldsymbol{X},Y)}\mathbb{E}_{\boldsymbol{M}}[l(f(\boldsymbol{M}\odot \boldsymbol{X};\boldsymbol{\theta}), Y)+\lambda ||\boldsymbol{M}||_0].
\end{equation}

In Equation \ref{eq:l0}, with the help of continuous probabilistic approximation, we can derive the $\ell_0$ norm of the mask as: $\mathbb{E}_{\boldsymbol{M}} ||\boldsymbol{M}||_0=\sum_{d=1}^{p} P(m_d>0)=\sum_{d=1}^{p} \Phi(\frac{\mu_d}{\sigma})$, where $\Phi$ is the cumulative distribution function of standard Gaussian. 
After learning sample weights corresponding to $\hat{w}(\boldsymbol{X})$, and combining them with the sparsity constraint, we rewrite Equation \ref{eq:l0} as below:

\begin{equation} \label{eq:stg}
\small
    \mathcal{L}(\boldsymbol{\theta}, \boldsymbol{\mu})=\mathbb{E}_{P(\boldsymbol{X},Y)}[\hat{w}(\boldsymbol{X})\mathbb{E}_{\boldsymbol{M}}[l(f(\boldsymbol{M}\odot \boldsymbol{X};\boldsymbol{\theta}), Y)+\lambda \sum_{d=1}^{p} \Phi(\frac{\mu_d}{\sigma})]]
\end{equation}

As a result, the optimization of Equation \ref{eq:stg} outputs model parameters $\boldsymbol{\theta}$, and soft masks $\boldsymbol{\mu}$ which are continuous variables in the range $[0, 1]$. 
Therefore, $\mu_d$ can be viewed as the probability of selecting $X_d$ as a stable variable. 
We may set a threshold to conduct variable selection, and then retrain a model only using selected variables for better covariate-shift generalization. 

\paragraph{Iterative procedure}
As mentioned before, global decorrelation between each pair of covariates could be too aggressive to accomplish. 
In reality, correlations inside stable variables could be strong. 
Global decorrelation may give rise to the shrinkage of effective sample size, causing inflated variance. 
It is worth noting that outputs of the backend module can be interpreted as $P(X_d\in \boldsymbol{S})$, namely the probability of each variable belonging to stable variables. They contain information on the covariate structure. 
Therefore, when using DWR as the frontend, we propose an approach to taking advantage of this information as feedback for the frontend module to mitigate the decrease in effective sample size. 

We first denote $\boldsymbol{A}\in [0,1]^{p\times p}$ as the covariance matrix mask, where $A_{ij}$ represents the strength of decorrelation for $X_i$ and $X_j$. 
Obviously, since we hope to preserve correlation inside stable variables $\boldsymbol{S}$, when the pair of variables is more likely to belong to $\boldsymbol{S}$, they should be less likely to be decorrelated. 
Thus the elements in $\boldsymbol{A}$ can be calculated as: $A_{ij}=1-P(X_i \in \boldsymbol{S})P(X_j \in \boldsymbol{S})=1-\mu_i \mu_j$.   
We incorporate this term into the loss function of DWR in Equation \ref{eq:DWR}, revising it as:

\begin{equation} \label{eq:maskDWR}
\small
    \mathcal{L}(\boldsymbol{w}) = \sum_{1\le i,j \le p, i\ne j}\left(A_{ij}Cov(X_i, X_j; \boldsymbol{w})\right)^2.
\end{equation}

Through Equation \ref{eq:maskDWR}, we implement SVI in an iterative way by combining sample reweighting and sparse learning. 
The details of this algorithm are described in Algorithm \ref{alg:iter}. We also present a diagram to illustrate it in Figure \ref{fig:iter}.

As we can see, when initialized, the frontend module $\mathcal{M}_{reweight}$ learns a group of sample weights corresponding to the weighting function $\hat{w}(\boldsymbol{X})$. 
Given such sample weights, the backend module $\mathcal{M}_{sparse}$ conducts sparse learning in a way of soft variable selection under the reweighted distribution, outputting the probability $P(X_d \in \boldsymbol{S})$ for each variable $X_d$ to be in the stable variable set. 
Such structural information can be utilized by $\mathcal{M}_{reweight}$ to learn better sample weights, since some of the correlations inside stable variables will be preserved. 
Therefore, sample reweighting and sparse learning modules benefit each other through such a loop of iteration and feedback. The iterative procedure and its convergence are hard to be analyzed theoretically, like in previous works \cite{liu2021heterogeneous, zhou2022model}, so we illustrate them through empirical experiments in Figure \ref{fig:period} and in appendix.

We denote Algorithm \ref{alg:iter} as \textbf{SVI} which applies to the linear setting since its frontend is DWR.
For nonlinear settings where DWR cannot address, we employ SRDO as the frontend module, denoted as \textbf{NonlinearSVI}. We do not implement it iteratively because it is hard for the resampling process of SRDO to incorporate the feedback of structural information from the backend module, which we leave for future extensions.
For both settings, unlike previous methods that directly learn a prediction model by optimizing the weighted loss, we first apply our algorithm to select the expected stable variables, then we retrain a model using these variables for prediction under covariate shift. 

\begin{algorithm}[t]
\caption{Sparse Variable Independence (SVI)} \label{alg:iter}
\begin{algorithmic}
    \STATE {\bfseries Input:} Dataset $[\boldsymbol{X}, \boldsymbol{Y}]$, sparse learning regularizer coefficient $\lambda$, number of loops $T$, period of moving average for soft mask $T_m$,  maximum number of iterations for learning weights $T_w$, maximum number of iterations for sparse learning $T_{\theta}$, selection threshold $\gamma$. 
    \STATE {\bfseries Output:} Selected variable set $\boldsymbol{S}'$. 
    \STATE Initialize $\boldsymbol{A}$ with $A_{ij}=1, \ \forall 1\leq i,j \leq p$
    \STATE Initialize $\boldsymbol{U}$ as empty list.
    \FOR {$t=1$ to $T$}
        \WHILE {not convergence or reach $T_w$} 
            \STATE Update sample weights $\boldsymbol{w}$ to minimize $\mathcal{L}(\boldsymbol{w})$ in Equation \ref{eq:maskDWR}.
        \ENDWHILE
        \WHILE {not convergence or reach $T_{\theta}$}
            \STATE Update model parameters $\boldsymbol{\theta}$ and soft mask probability $\boldsymbol{\mu}$ to minimize $\mathcal{L}(\boldsymbol{\theta}, \boldsymbol{\mu})$ in Equation \ref{eq:stg}.
        \ENDWHILE
        \STATE Delete first element of $\boldsymbol{U}$ if its length is greater than $T_m$.
        \STATE Append current $\boldsymbol{\mu}$ to the end of $\boldsymbol{U}$.
        \STATE Calculate the moving average $\boldsymbol{\mu}'$ from $\boldsymbol{U}$.
        \STATE Calculate $A_{ij}=1-\mu_i' \mu_j', \ \forall 1\leq i,j \leq p$.
    \ENDFOR
    \STATE Calculate $\boldsymbol{S}'=\{X_d \ | \ \mu_d \geq \gamma \}$
    \STATE {\bfseries return: $\boldsymbol{S}'$} 
\end{algorithmic}
\end{algorithm}

\section{Experiments}

\begin{figure*}[t]
	\centering
	\subfigure[RMSE of linear settings when fixing $r_{train}=2.5$ with varying $n$. ] {
	\label{fig:samplesize}
	    \includegraphics[width=0.32\linewidth]{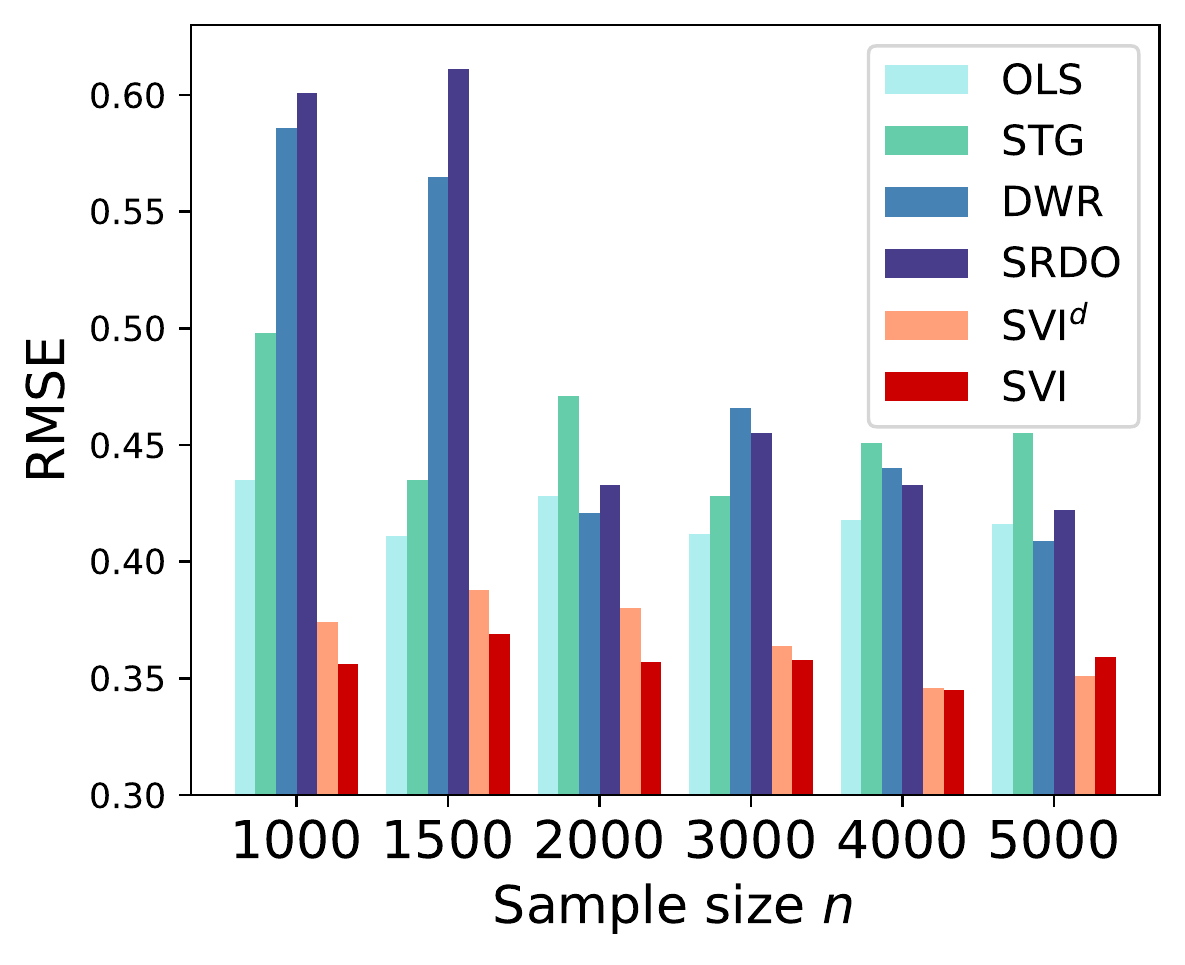}
	}
	\subfigure[RMSE of the linear setting with bias rate $r_{train}=2.5$ and sample size $n=2000$. ] {
	\label{fig:linear}
	    \includegraphics[width=0.32\linewidth]{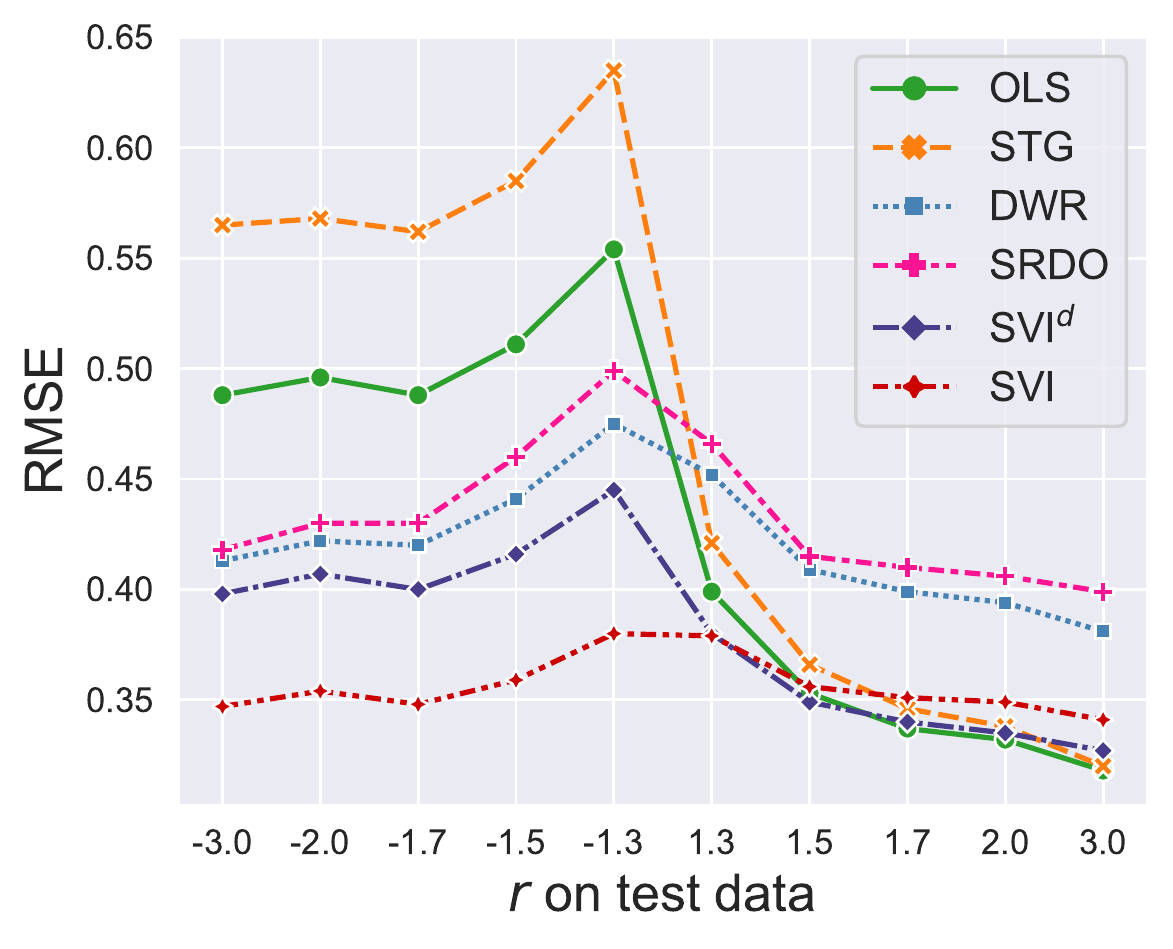}
	}
	\subfigure[RMSE of the nonlinear setting with bias rate $r_{train}=1.8$ and sample size $n=25000$. ]   {
	\label{fig:nonlinear}
	    \includegraphics[width=0.32\linewidth]{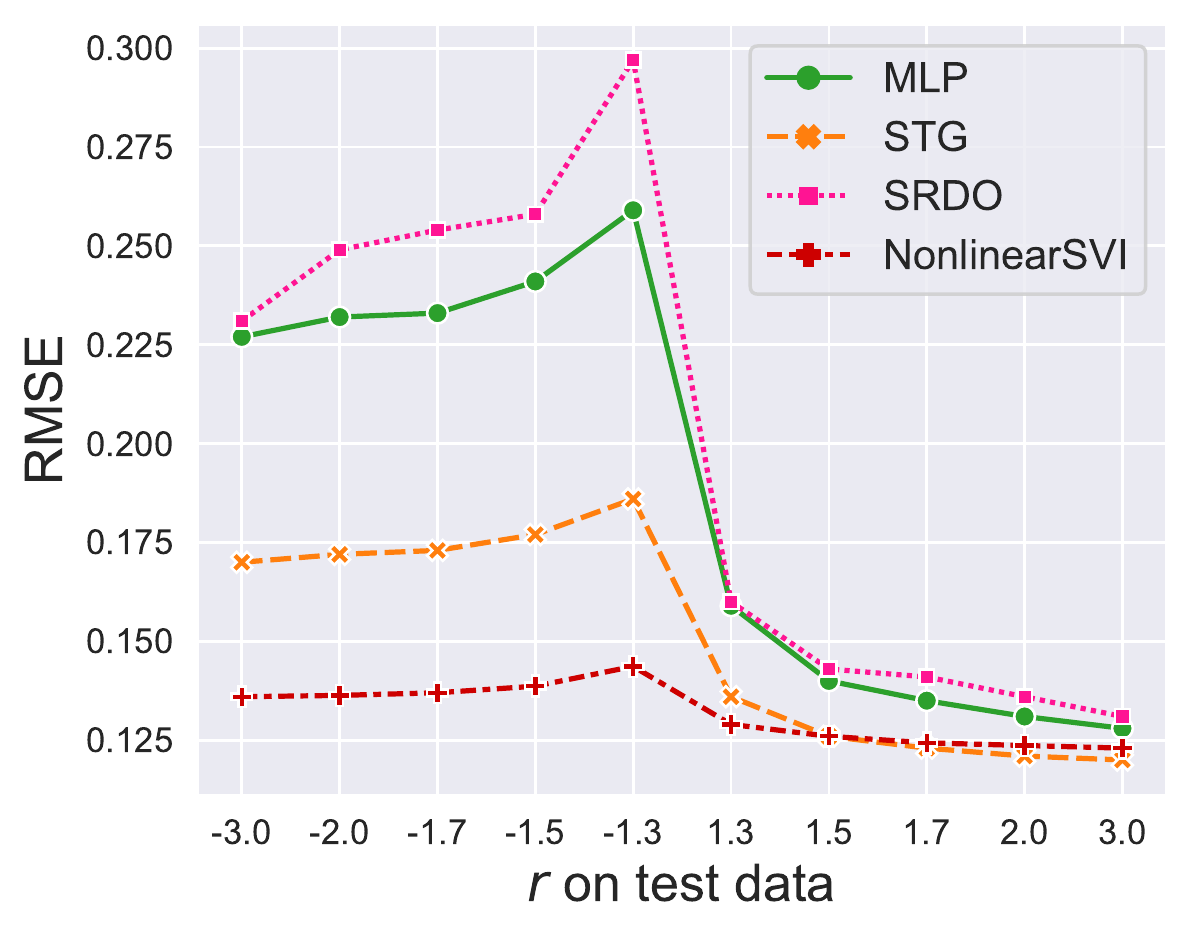}
	}
	\caption{Detailed results of experiments on synthetic data under linear and nonlinear settings. }
	\label{fig:sim}
\end{figure*}

% \usepackage{booktabs}
\begin{table*}[!ht]
  \centering
  \caption{Results of the linear setting under varying sample size $n$ and train data bias rate $r$. }
  \label{table:sim-linear}
\resizebox{0.9\linewidth}{!}{
\begin{tabular}{@{}cccccccccc@{}}
\toprule
\multicolumn{10}{c}{Scenario 1: Varying sample size $n (r=2.5)$}                                                                                                                                                      \\ \midrule
\multicolumn{1}{c|}{$n$}       & \multicolumn{3}{c|}{$n=1000$}                                           & \multicolumn{3}{c|}{$n=1500$}                                           & \multicolumn{3}{c}{$n=2000$}                       \\ \midrule
\multicolumn{1}{c|}{Methods} & Mean\_Error    & Std\_Error     & \multicolumn{1}{c|}{Max\_Error}     & Mean\_Error    & Std\_Error     & \multicolumn{1}{c|}{Max\_Error}     & Mean\_Error    & Std\_Error     & Max\_Error     \\ \midrule
\multicolumn{1}{c|}{OLS}     & 0.435          & 0.084          & \multicolumn{1}{c|}{0.572}          & 0.411          & 0.072          & \multicolumn{1}{c|}{0.531}          & 0.428          & 0.084          & 0.554          \\
\multicolumn{1}{c|}{STG}     & 0.498          & 0.133          & \multicolumn{1}{c|}{0.694}          & 0.435          & 0.090           & \multicolumn{1}{c|}{0.579}          & 0.471          & 0.116          & 0.635          \\
\multicolumn{1}{c|}{DWR}     & 0.586          & 0.043          & \multicolumn{1}{c|}{0.676}          & 0.565          & 0.044          & \multicolumn{1}{c|}{0.666}          & 0.421          & 0.027          & 0.475          \\ 
\multicolumn{1}{c|}{SRDO}     & 0.601          & 0.040          & \multicolumn{1}{c|}{0.698}          & 0.611          & 0.045           & \multicolumn{1}{c|}{0.690}          & 0.433          & 0.035          & 0.499          \\ \midrule
\multicolumn{1}{c|}{SVI$^d$}     & 0.374          & 0.023          & \multicolumn{1}{c|}{\textbf{0.413}} & 0.388          & 0.043          & \multicolumn{1}{c|}{0.465}          & 0.380           & 0.038          & 0.445          \\
\multicolumn{1}{c|}{SVI}    & \textbf{0.356} & \textbf{0.020}  & \multicolumn{1}{c|}{0.415}          & \textbf{0.369} & \textbf{0.030}  & \multicolumn{1}{c|}{\textbf{0.430}}  & \textbf{0.357} & \textbf{0.013} & \textbf{0.380}  \\ \midrule
\multicolumn{10}{c}{Scenario 2: Varying bias rate $r (n=1500)$}                                                                                                                                                     \\ \midrule
\multicolumn{1}{c|}{$r$}       & \multicolumn{3}{c|}{$r=2.0$}                                            & \multicolumn{3}{c|}{$r=2.5$}                                            & \multicolumn{3}{c}{$r=3.0$}                        \\ \midrule
\multicolumn{1}{c|}{Methods} & Mean\_Error    & Std\_Error     & \multicolumn{1}{c|}{Max\_Error}     & Mean\_Error    & Std\_Error     & \multicolumn{1}{c|}{Max\_Error}     & Mean\_Error    & Std\_Error     & Max\_Error     \\ \midrule
\multicolumn{1}{c|}{OLS}     & 0.374          & 0.039          & \multicolumn{1}{c|}{0.453}          & 0.411          & 0.072          & \multicolumn{1}{c|}{0.531}          & 0.443          & 0.100            & 0.589          \\
\multicolumn{1}{c|}{STG}     & 0.395          & 0.047          & \multicolumn{1}{c|}{0.475}          & 0.435          & 0.090           & \multicolumn{1}{c|}{0.579}          & 0.526          & 0.165          & 0.749          \\
\multicolumn{1}{c|}{DWR}     & 0.431          & 0.031          & \multicolumn{1}{c|}{0.499}          & 0.565          & 0.044          & \multicolumn{1}{c|}{0.666}          & 0.576          & 0.038          & 0.655          \\ 
\multicolumn{1}{c|}{SRDO}     & 0.457          & 0.058          & \multicolumn{1}{c|}{0.511}          & 0.611          & 0.045           & \multicolumn{1}{c|}{0.690}          & 0.592          & 0.037          & 0.643          \\ \midrule
\multicolumn{1}{c|}{SVI$^d$}     & 0.356          & 0.017          & \multicolumn{1}{c|}{0.395}          & 0.388          & 0.043          & \multicolumn{1}{c|}{0.465}          & 0.364          & 0.015          & 0.393          \\
\multicolumn{1}{c|}{SVI}    & \textbf{0.345} & \textbf{0.015} & \multicolumn{1}{c|}{\textbf{0.380}}  & \textbf{0.369} & \textbf{0.030}  & \multicolumn{1}{c|}{\textbf{0.430}}  & \textbf{0.330}  & \textbf{0.014} & \textbf{0.377} \\ \bottomrule
\end{tabular}
}
\end{table*}

% \usepackage{booktabs}
\begin{table*}[!ht]
  \centering
  \caption{Results of the nonlinear setting under varying sample size $n$ and train data bias rate $r$. }
  \label{table:sim-nonlinear}
\resizebox{0.9\linewidth}{!}{
  \begin{tabular}{@{}cccccccccc@{}}
\toprule
\multicolumn{10}{c}{Scenario 1: Varying sample size $n (r=2.0)$}                                                                                                                                                           \\ \midrule
\multicolumn{1}{c|}{$n$}          & \multicolumn{3}{c|}{$n$=15000}                                        & \multicolumn{3}{c|}{$n=20000$}                                        & \multicolumn{3}{c}{$n=25000$}                    \\ \midrule
\multicolumn{1}{c|}{Methods}      & Mean\_Error    & Std\_Error     & \multicolumn{1}{c|}{Max\_Error}     & Mean\_Error    & Std\_Error     & \multicolumn{1}{c|}{Max\_Error}     & Mean\_Error    & Std\_Error     & Max\_Error     \\ \midrule
\multicolumn{1}{c|}{MLP}          & 0.221          & 0.080          & \multicolumn{1}{c|}{0.331}          & 0.262          & 0.113          & \multicolumn{1}{c|}{0.416}          & 0.249          & 0.104          & 0.389          \\
\multicolumn{1}{c|}{STG}          & 0.177          & 0.049          & \multicolumn{1}{c|}{0.243}          & 0.176          & 0.048          & \multicolumn{1}{c|}{0.241}          & 0.176          & 0.048          & 0.243          \\ 
\multicolumn{1}{c|}{SRDO}          & 0.244          & 0.123          & \multicolumn{1}{c|}{0.380}          & 0.288          & 0.133          & \multicolumn{1}{c|}{0.469}          & 0.231          & 0.090          & 0.373          \\ \midrule
\multicolumn{1}{c|}{NonlinearSVI} & \textbf{0.130} & \textbf{0.002} & \multicolumn{1}{c|}{\textbf{0.133}} & \textbf{0.125} & \textbf{0.001} & \multicolumn{1}{c|}{\textbf{0.128}} & \textbf{0.126} & \textbf{0.002} & \textbf{0.129} \\ \midrule
\multicolumn{10}{c}{Scenario 2: Varying bias rate $r (n=25000)$}                                                                                                                                                            \\ \midrule
\multicolumn{1}{c|}{$r$}          & \multicolumn{3}{c|}{$r=1.8$}                                          & \multicolumn{3}{c|}{$r=2.0$}                                          & \multicolumn{3}{c}{$r=2.2$}                      \\ \midrule
\multicolumn{1}{c|}{Methods}      & Mean\_Error    & Std\_Error     & \multicolumn{1}{c|}{Max\_Error}     & Mean\_Error    & Std\_Error     & \multicolumn{1}{c|}{Max\_Error}     & Mean\_Error    & Std\_Error     & Max\_Error     \\ \midrule
\multicolumn{1}{c|}{MLP}          & 0.188          & 0.051          & \multicolumn{1}{c|}{0.259}          & 0.249          & 0.104          & \multicolumn{1}{c|}{0.389}          & 0.498          & 0.312          & 0.901          \\
\multicolumn{1}{c|}{STG}          & 0.150          & 0.026          & \multicolumn{1}{c|}{0.186}          & 0.176          & 0.048          & \multicolumn{1}{c|}{0.243}          & 0.208          & 0.076          & 0.308          \\ 
\multicolumn{1}{c|}{SRDO}          & 0.200          & 0.071          & \multicolumn{1}{c|}{0.297}          & 0.236          & 0.089          & \multicolumn{1}{c|}{0.353}          & 0.469          & 0.203          & 0.717          \\ \midrule
\multicolumn{1}{c|}{NonlinearSVI} & \textbf{0.132} & \textbf{0.007} & \multicolumn{1}{c|}{\textbf{0.144}} & \textbf{0.126} & \textbf{0.002} & \multicolumn{1}{c|}{\textbf{0.129}} & \textbf{0.126} & \textbf{0.002} & \textbf{0.129} \\ \bottomrule
\end{tabular}
}
\end{table*}

% In this section, we examine the covariate-shift generalization ability of our algorithms with experiments on both synthetic and real-world datasets. 

\begin{figure*}[t]
	\centering
	\subfigure[$||\beta_v||_1$ with varying sample size $n$. ] {
	\label{fig:betav}
	    \includegraphics[width=0.32\linewidth]{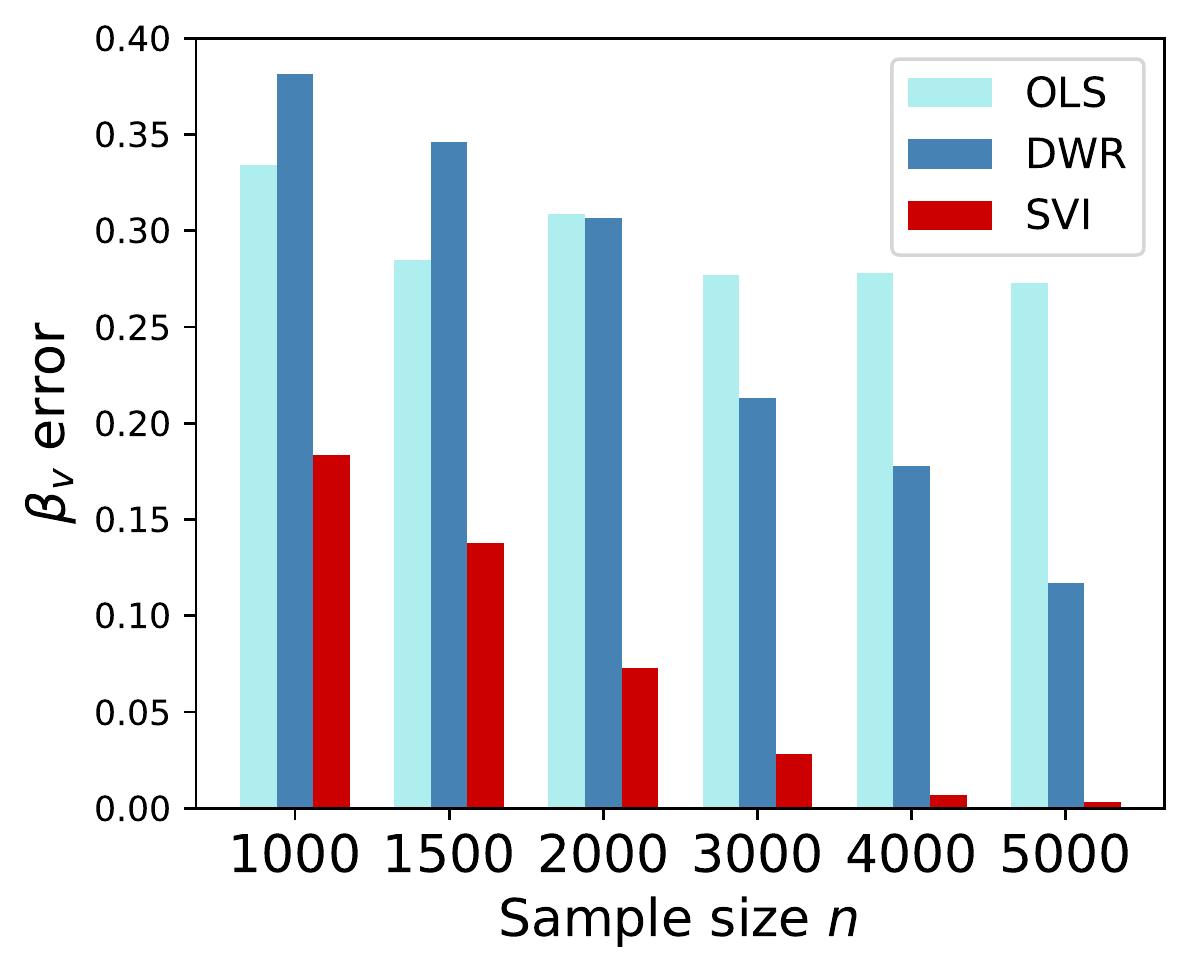}
	}
	\subfigure[$n_{eff} $ with varying sample size $n$. ] {
	\label{fig:neff}
	    \includegraphics[width=0.32\linewidth]{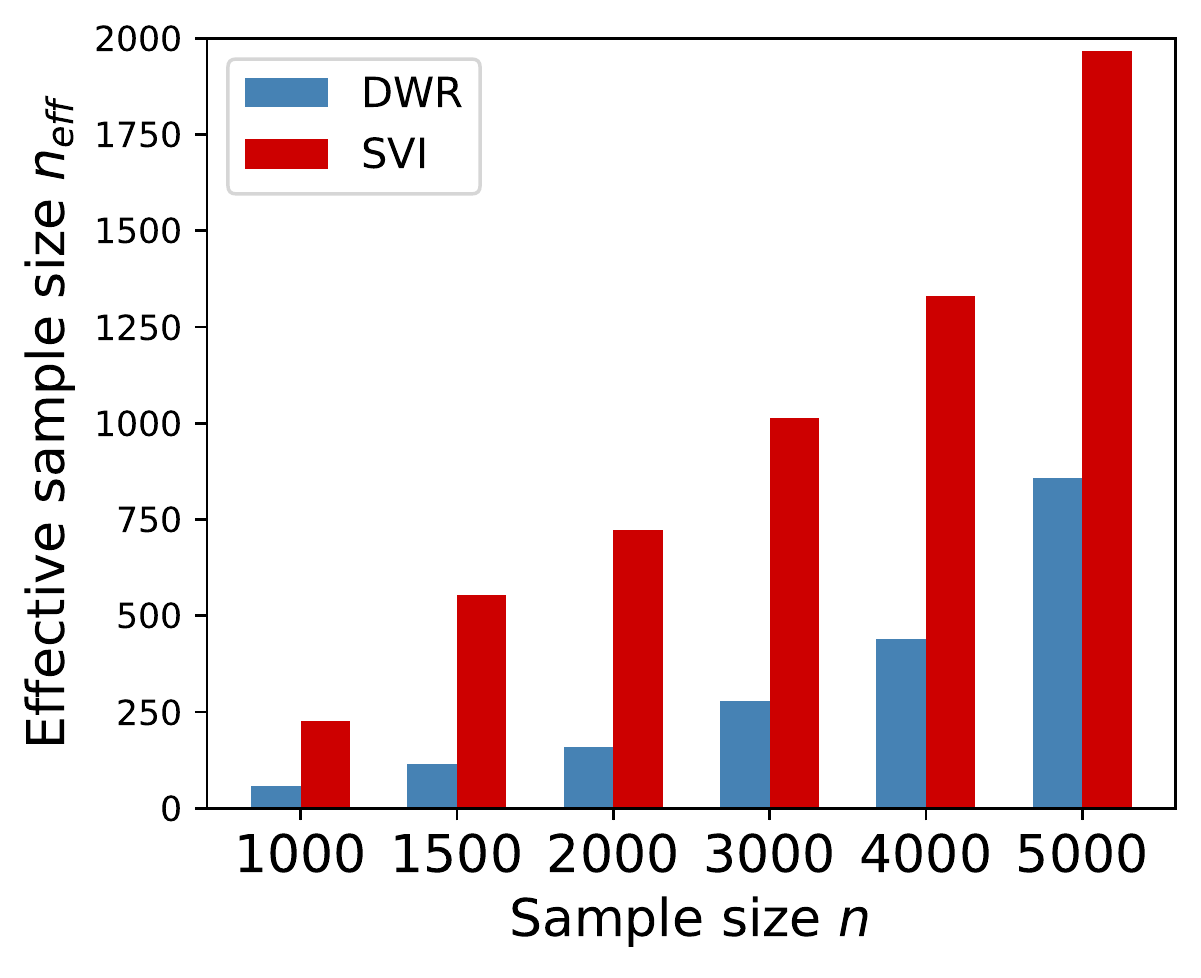}
	}
	\subfigure[$||\beta_v||_1$ with growing \#iterations. ] {
	\label{fig:period}
	    \includegraphics[width=0.32\linewidth]{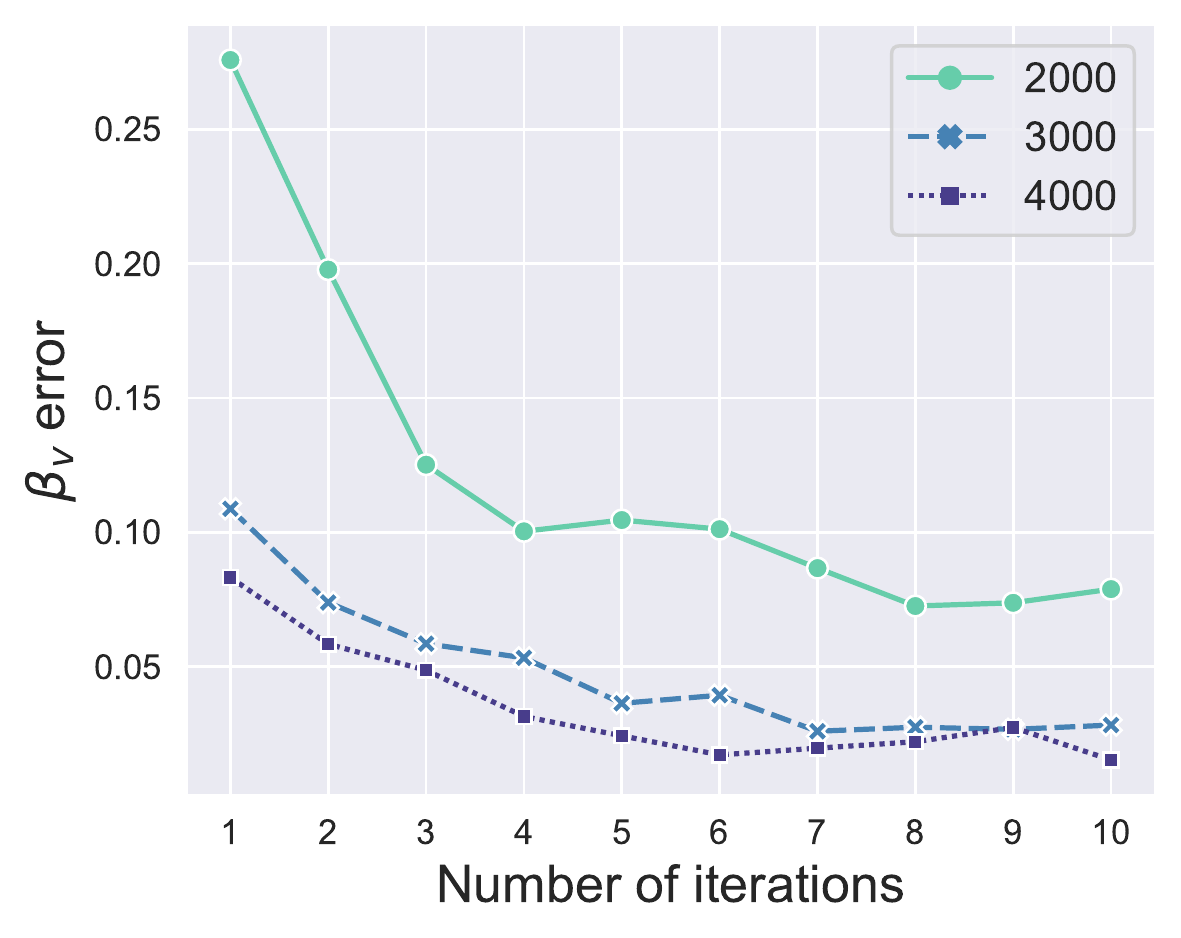}
	}
	\caption{Additional results for illustrating effectiveness of SVI, when fixing $r_{train}=2.5$. }
	\label{fig:addition}
\end{figure*}

\subsection{Baselines}

We compare SVI with the following methods. We tune the hyperparameters by grid search and validation on data from the environment corresponding to $r_{train}$. 
\begin{itemize}
    \item OLS (Ordinary Least Squares):  For linear settings. 
    \item MLP (Multi-Layer Perceptron): For nonlinear settings. 
    \item STG (Stochastic Gates) \cite{yamada2020feature}: Directly optimizing Equation \ref{eq:stg} without sample reweighting.
    \item DWR \cite{kuang2020stable}: Optimizing Equation \ref{eq:DWR} and conducting Weighted Least Squares (for linear settings). 
    \item SRDO \cite{shen2020stable}: Conducting density ratio estimation through Equation \ref{eq:SRDO}. 
    \item SVI$^d$: A degenerated version of SVI by running only one iteration for ablation study to demonstrate the benefit of iterative procedure under linear settings.
\end{itemize}

\subsection{Evaluation Metrics}
To evaluate the covariate-shift generalization performance and stability, we assess the algorithms on test data from multiple different environments, thus we adopt the following metrics: $Mean\_Error = \frac{1}{|\varepsilon_{te}|}\sum_{e\in \varepsilon_{te}}\mathcal{L}^e$;
    $Std\_Error = \sqrt{\frac{1}{|\varepsilon_{te}|-1} \sum_{e\in \varepsilon_{te}}(\mathcal{L}^e-Mean\_Error)^2}$;
    $Max\_Error = \max_{e\in \varepsilon_{te}}\mathcal{L}^e$;
$\varepsilon_{te}$ denotes the testing environments, $\mathcal{L}^e$ denotes the empirical loss in the test environment $e$.

\subsection{Experiments on Synthetic Data}
\subsubsection{Dataset}
We generate $\boldsymbol{X}=\{\boldsymbol{S}, \boldsymbol{V}\}$ from a multivariate Gaussian distribution $\boldsymbol{X}\sim N(\boldsymbol{0}, \boldsymbol{\Sigma})$.
In this way, we can simulate different correlation structures of $\boldsymbol{X}$ through the control of covariance matrix $\boldsymbol{\Sigma}$. 
In our experiments, we tend to make the correlations inside stable variables strong.
% For simplicity, We further set $\boldsymbol{\Sigma}=Diag \left( \boldsymbol{\Sigma}^{(S)}, \boldsymbol{\Sigma}^{(V)} \right)$, assuming block structure. 
% For $\boldsymbol{\Sigma}^{(S)} \in \mathbb{R}^{p_s \times p_s}$, its elements are: $\boldsymbol{\Sigma}^{(S)}_{jk}=\rho_s$ for $j \neq k$ and $\boldsymbol{\Sigma}^{(S)}_{jk}=1$ for $j=k$. 
% For $\boldsymbol{\Sigma}^{(V)} \in \mathbb{R}^{p_v \times p_v}$, its elements are: $\boldsymbol{\Sigma}^{(V)}_{jk}=\rho_v$ for $j \neq k$ and $\boldsymbol{\Sigma}^{(V)}_{jk}=1$ for $j=k$. 
% Despite we can simulate more complex settings by manipulating the covariance matrix $\boldsymbol{\Sigma}$, such a simplified setting is enough since it can adjust correlation inside $\boldsymbol{S}$ or inside $\boldsymbol{V}$. 
% In our experiments, to simulate the scenario where correlations inside stable variables are strong, we tend to set $\rho_s$ a large value. We set $\rho_s=0.9$ and $\rho_v=0.1$ in our experiments.

For linear and nonlinear settings, we employ different data generation functions. It is worth noting that a certain degree of model misspecification is needed, otherwise the model will be able to directly learn the true stable variables simply using OLS or MLP. 

For the linear setting, we introduce the model misspecification error by adding an extra polynomial term to the dominated linear term, and later use a linear model to fit the data. The generation function is as follows:
\begin{equation}
\small
    Y=f(\boldsymbol{S})+\epsilon=[\boldsymbol{S}, \boldsymbol{V}]\cdot [\boldsymbol{\beta}_s, \boldsymbol{\beta}_v]^T + \boldsymbol{S}_{\cdot, 1}\boldsymbol{S}_{\cdot, 2}\boldsymbol{S}_{\cdot, 3} + \epsilon
\end{equation}
% where $\boldsymbol{\beta}_s=\{\frac{1}{3}, -\frac{2}{3}, 1, -\frac{1}{3}, \frac{2}{3}, -1,... \}, \boldsymbol{\beta}_v=\boldsymbol{0}, \epsilon \sim \mathcal{N}(0, 0.3)$. 

For the nonlinear setting, we generate data in a totally nonlinear fashion. We employ random initialized MLP as the data generation function.
\begin{equation}
\small
    Y=f(\boldsymbol{S})+\epsilon = MLP(\boldsymbol{S})+\epsilon 
\end{equation}
Later we use MLP with a smaller capacity to fit the data. 
% where the MLP is 2-layer with 100 hidden units, $\epsilon\sim \mathcal{N}(0, 0.1)$. To introduce model misspecification, we use a 2-layer MLP with 10 hidden units to fit the data after employing SVI to do variable selection. Parameters of MLP are initialized with uniform distribution $U(-1,1)$. 
More details are included in appendix. 

\subsubsection{Generating Various Environments}
To simulate the scenario of covariate shift and test not only the prediction accuracy but prediction stability, we generate a set of environments each with a distinct distribution.
Specifically, following \cite{shen2020stable2}, we generate different environments in our experiments through changing $P(\boldsymbol{V}|\boldsymbol{S})$, further leading to the change of $P(Y|\boldsymbol{V})$. 
Among all the unstable variables $\boldsymbol{V}$, we simulate unstable correlation $P(\boldsymbol{V}_b|\boldsymbol{S})$ on a subset $\boldsymbol{V}_b\in \boldsymbol{V}$.
% where the dimension of $\boldsymbol{V}_b$ can be tuned. 
We vary $P(\boldsymbol{V}_b|\boldsymbol{S})$ through different strengths of selection bias with a bias rate $r\in[-3, -1)\cup(1,3]$. 
For each sample, we select it with probability $Pr=\Pi_{V_i \in \boldsymbol{V}_b}|r|^{-5D_i}$, where $D_i=|f(\boldsymbol{S})-sign(r)V_i|$, $sign$ denotes sign function.
In our experiments, we set $p_{v_b}=0.1*p$.

% We can find that $r>1$ corresponds to positive correlation between $Y$ and $\boldsymbol{V}_b$, and $r<-1$ refers to negative correlation between $Y$ and $\boldsymbol{V}_b$. 
% A larger value of $|r|$ implies a stronger correlation between $\boldsymbol{V}_b$ and $Y$. By varying the bias rate $r$, we can simulate different environments. 

\begin{figure*}[t]
	\centering
	\subfigure[Prediction error for house price] {
	\label{fig:house}
	    \includegraphics[width=0.4\linewidth]{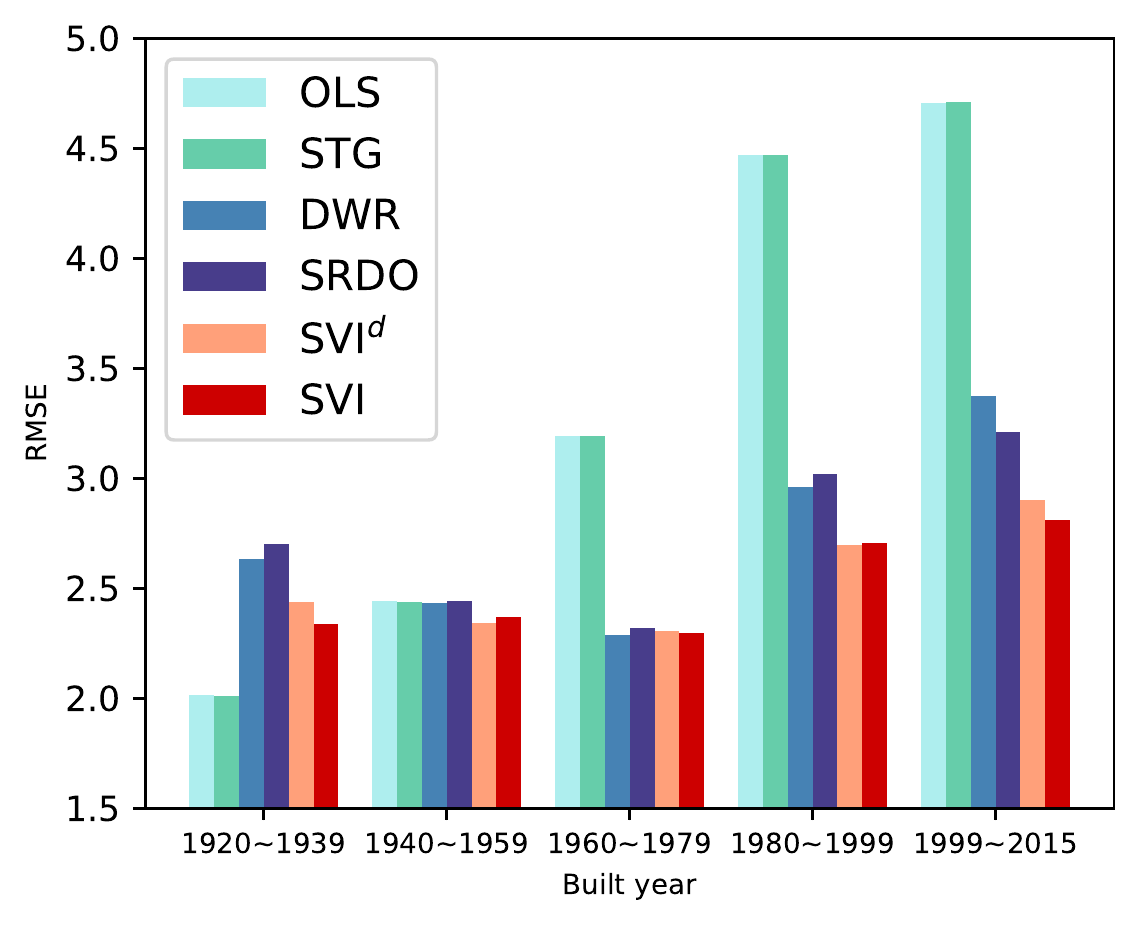}
	}
	\subfigure[Classification error for people income prediction] {
	\label{fig:adult}
	    \includegraphics[width=0.4\linewidth]{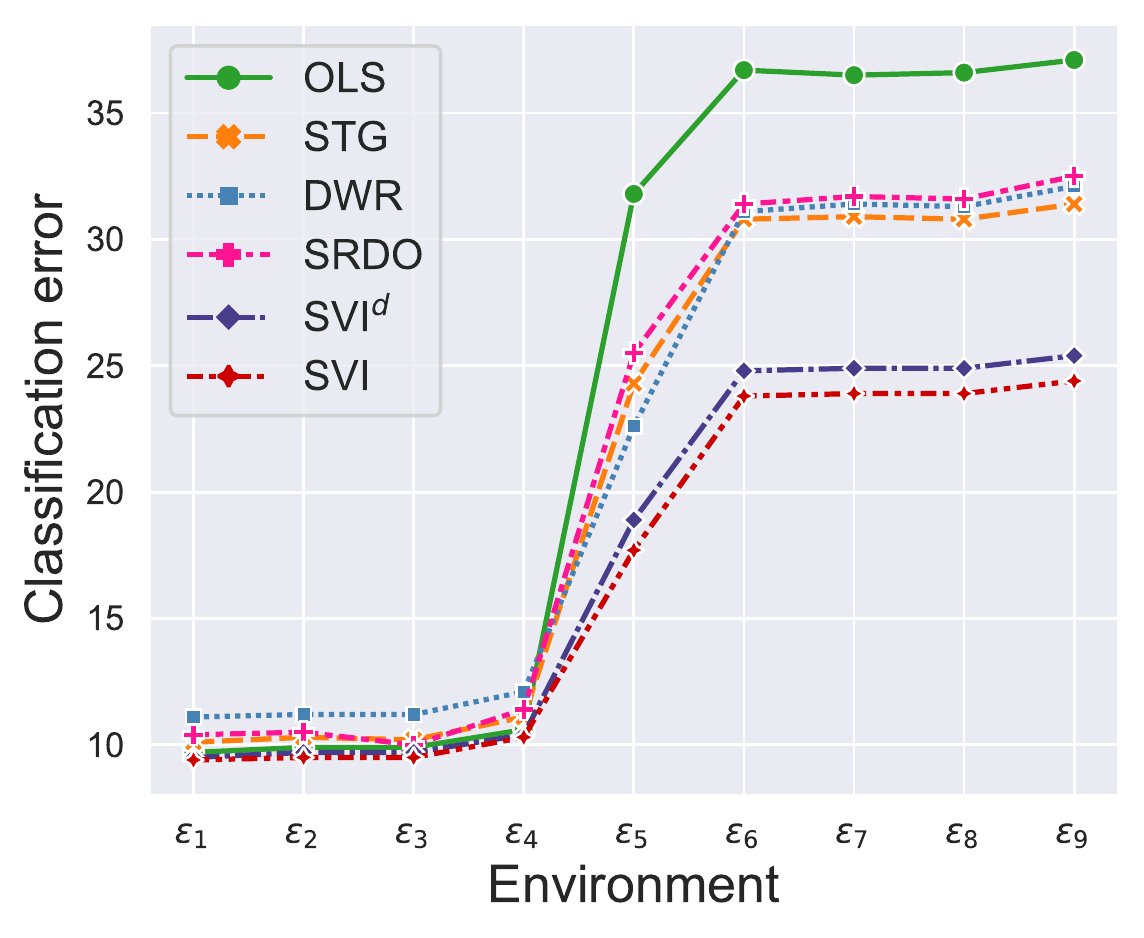}
	}
	\caption{Detailed results of experiments on real-world data. }
	\label{fig:real}
\end{figure*}

\subsubsection{Experimental settings}
% In our experiments, we evaluate the performance of all algorithms via the three metrics: Mean\_Error, Std\_Error, and Max\_Error. 
We train our models on data from one single environment generated with a bias rate $r_{train}$ and test on data from multiple environments with bias rates $r_{test}$ ranging in $[-3, -1)\cup(1,3]$. 
Each model is trained 10 times independently with different training datasets from the same bias rate $r_{train}$. 
Similarly, for each $r_{test}$, we generate 10 different test datasets. The metrics we report are the mean results of these 10 times. 

\subsubsection{Results}
Results are shown in Table \ref{table:sim-linear} and \ref{table:sim-nonlinear} when varying sample size $n$ and training data bias rate $r_{train}$. 
Detailed results for two specific settings are illustrated in Figure \ref{fig:linear} and \ref{fig:nonlinear}. 
In addition to prediction performance, we illustrate the effectiveness of our algorithm in weakening residual correlations and increasing effective sample size in Figure \ref{fig:betav} and \ref{fig:neff}. 
Analysis of the results is as follows:

\begin{itemize}
    \item From Table \ref{table:sim-linear} and \ref{table:sim-nonlinear}, for almost every setting, SVI consistently outperforms other baselines in Mean\_Error, Std\_Error, and Max\_Error, indicating its superior covariate-shift generalization ability and stability. From Figure \ref{fig:linear} and \ref{fig:nonlinear}, when $r_{test} < -1$, i.e. the correlation between $\boldsymbol{V}$ and $Y$ reverses in test data compared with that in training data, other baselines fail significantly while SVI remains stable against such challenging distribution shift. 
    \item From Figure \ref{fig:samplesize}, we find that there is a sharp rise of prediction error for DWR and SRDO with the decrease of sample size, confirming the severity of the over-reduced effective sample size. Meanwhile, SVI maintains great performance and generally outperforms SVI$^d$, demonstrating the superiority of the iteration procedure which avoids aggressive decorrelation within $\boldsymbol{S}$. 
    \item In Figure \ref{fig:betav}, we calculate $||\beta_v||_1$ to measure the residual correlations between unstable variables $\boldsymbol{V}$ and the outcome $Y$. 
    % For SVI, we calculate $\beta_v$ by multiplying the model parameters and the soft masks. 
    DWR always preserves significant non-zero coefficients on unstable variables especially when the sample size gets smaller, while SVI truncates the scale of residual correlations by one or two orders of magnitude sometimes. This strongly illustrates that SVI indeed helps alleviate the imperfectness of variable decorrelation in previous independence-based sample reweighting methods. 
    \item In Figure \ref{fig:neff}, we calculate $n_{eff}=\frac{(\sum_{i=1}^n w_i)^2}{\sum_{i=1}^n w_i^2}$ as effective sample size following \cite{kish1965survey}. It is evident that SVI greatly boosts the effective sample size compared with global decorrelation schemes like DWR, since there are strong correlations inside stable variables. The shrinkage of $n_{eff}$ compared with the original sample size $n$ becomes rather severe for DWR when $n$ is relatively small, even reaching 1/10. 
    \item In Figure \ref{fig:period} we plot the change of $||\beta_v||_1$ with the increasing number of iterations of SVI. We can observe that the residual correlations decline as the algorithm evolves. This proves the benefit brought by our iterative procedure. Also, this demonstrates the numerical convergence of SVI since coefficients of unstable variables $\boldsymbol{V}$ gradually approach zero with the iterative process. More experiments for analyzing the iterative process and convergence are included in appendix.
\end{itemize}

\subsection{Experiments on Real-World Data}
\subsubsection{House Price Prediction}

% This task comes from a real-world dataset of houses from King County, USA. We acquire this dataset from Kaggle\footnote{\url{https://www.kaggle.com/c/house-prices-advanced-regression-techniques/data}}. 
% It is a regression task, where the outcome is the house price and each sample contains 20 variables as potential inputs, such as the number of bedrooms or bathrooms. We first drop 3 useless columns like zip-code. Then we simulate different environments according to the built year, dividing the dataset into 6 periods. Each period is approximately two decades. To examine the covariate-shift generalization ability, we train on the first environment in which houses were built between 1900 and 1919, and test on the other 5 periods. 

It is a regression task for predicting house price, splitting data into 6 environments, 1 for training and 5 for testing, according to the time period that the house was built.\footnote{Due to the limited space, a detailed description of the two real-world datasets is presented in appendix.}
In 4 out of 5 test environments, SVI and SVI$^d$ outperform other baselines, especially in the last 3 environments. The gap significantly increases along the time axis, which represents a longer time span between test data and training data. This implies that a severer distribution shift may embody the superiority of our algorithm. Besides, overall SVI performs slightly better than SVI$^d$, further demonstrating the benefit of iterative procedure on real-world data.

\subsubsection{People Income Prediction}
It is a binary classification task for income prediction where 10 environments are generated by the combination of race and sex. 
% In this experiment we employ the Adult dataset \cite{kohavi1996scaling}, formulating a binary classification task. The target variable is whether an adult's income per year exceeds 50K. We notice that the variables race and sex naturally produce sub-populations in the dataset, thus we use a combination of race and sex to generate 10 environments. 
We train the models on the first environment $(White, Female)$ and test on the other 9 environments to simulate distribution shift. 

For the first 4 environments where people's gender remains the same as that in training data, i.e. both are female, these methods make little difference in prediction. However, for the last 5 environments, when the sex category is male, performance drops for every method with varying degrees. We can see that SVI methods are the most stable ones, whose performance is affected much less than other baselines in the presence of distribution shift. Moreover, SVI still outperforms SVI$^d$ slightly, indicating the practical use of the iteration procedure again.

\section{Conclusion}

In this paper, we combined sample reweighting and sparsity constraint to compensate for the deficiency of independence-based sample reweighting when there is a residual dependency between stable and unstable variables, and the variance inflation caused by removing strong correlations inside stable variables. Experiments on both synthetic datasets and real-world datasets demonstrated the effectiveness of our algorithm. 

\section{Acknowledgements}

This work was supported in part by National Key R\&D Program of China (No. 2018AAA0102004), National Natural Science Foundation of China (No. U1936219, 62141607), Beijing Academy of Artificial Intelligence (BAAI). 
Peng Cui is the corresponding author. 
All opinions in this paper are those of the authors and do not necessarily reflect the views of the funding agencies.

\appendix

\section{Appendix}

\subsection{Proofs}
To differentiate between notations of sample and population, for notations of sample level, we add a superscript of $(n)$, while for notations of population level we do not. 
For example, $\boldsymbol{X}$ denotes a $p\times 1$ multidimensional random vector, while $\boldsymbol{X}^{(n)}$ denotes a $n\times p$ data matrix. ${\rm Cov}^{(n)}$ denotes sample covariance matrix while ${\rm Cov}$ denotes covariance matrix of population.

We need a lemma from \cite{zhao2006model} to prove Theorem \ref{prop:lasso}. The lemma is stated below.
\begin{lemma}
\label{lemma:lasso}
    Assume Equation \ref{eq:cov} holds. then we have
\begin{equation}
    P(\hat{\boldsymbol{\beta}}(\lambda_n)=_s \boldsymbol{\beta})\geq P(A\cap B).
\end{equation}
for
\begin{equation}
\begin{aligned}
A&=\{|({\rm Cov}^{(n)}(\boldsymbol{S}))^{-1}\boldsymbol{W}_s^{(n)}|\\
&<\sqrt{n}(|\boldsymbol{\beta_s}|-\frac{\lambda_n}{2n}|({\rm Cov}^{(n)}(\boldsymbol{S}))^{-1}{\rm sign}(\boldsymbol{\beta}_s)|)\}. \\
B&=\{|{\rm Cov}^{(n)}(\boldsymbol{V}, \boldsymbol{S})({\rm Cov}^{(n)}(\boldsymbol{S}))^{-1}\boldsymbol{W}_s^{(n)}-\boldsymbol{W}_v^{(n)}|\\
&\leq \frac{\lambda_n}{2\sqrt{n}}\boldsymbol{\eta}\}. \\ \nonumber
\end{aligned}
\end{equation}
where
\begin{equation}
\begin{aligned}
\boldsymbol{W}^{(n)}&=(\boldsymbol{X}^{(n)})^T(g^{(n)}(\boldsymbol{S})+\boldsymbol{\epsilon}^{(n)})/\sqrt{n}. \\
\boldsymbol{W}_s^{(n)}&=(\boldsymbol{S}^{(n)})^T(g^{(n)}(\boldsymbol{S})+\boldsymbol{\epsilon}^{(n)})/\sqrt{n}. \\
\boldsymbol{W}_v^{(n)}&=(\boldsymbol{V}^{(n)})^T(g^{(n)}(\boldsymbol{S})+\boldsymbol{\epsilon}^{(n)})/\sqrt{n}. \\ \nonumber
\end{aligned}
\end{equation}
\end{lemma}

\textbf{Proof of Theorem \ref{prop:lasso}}. By Lemma \ref{lemma:lasso}, we have
\begin{equation}
P(\hat{\boldsymbol{\beta}}(\lambda_n)=_s \boldsymbol{\beta})\geq P(A\cap B).
\end{equation}
where 
\begin{equation}
\begin{aligned}
    &\enspace \quad 1-P(A\cap B) \\
    &\leq P(A)+P(B) \\ 
    &\leq \sum_{i=1}^{p_s}P(|\zeta_i^{(n)}|\geq \sqrt{n}(|\beta_i|-\frac{\lambda_n}{2n}b_i^{(n)})\\
    &+\sum_{i=1}^{p_v}P(|\xi_i^{(n)}|\geq \frac{\lambda_n}{2\sqrt{n}}\eta_i) \\ \nonumber
\end{aligned}
\end{equation}

where $\boldsymbol{\zeta}^{(n)}=({\rm Cov}^{(n)}(\boldsymbol{S}))^{-1}\boldsymbol{W}_s^{(n)}, \boldsymbol{\xi}^{(n)}={\rm Cov}^{(n)}(\boldsymbol{V}, \boldsymbol{S})({\rm Cov}^{(n)}(\boldsymbol{S}))^{-1}\boldsymbol{W}_s^{(n)}-\boldsymbol{W}_v^{(n)}$, $\boldsymbol{b}^{(n)}=({\rm Cov}^{(n)}(\boldsymbol{S}))^{-1}{\rm sign}(\boldsymbol{\beta}_s)$. 

According to the weak law of large numbers, we have ${\rm Cov}^{(n)}(\boldsymbol{S}) \rightarrow_p {\rm Cov}(\boldsymbol{S})$, where $\rightarrow_p$ denotes converge in probability. 
Since the inverse of a matrix is a continuous function, according to the continuous mapping theorem, we have
$({\rm Cov}^{(n)}(\boldsymbol{S}))^{-1} \rightarrow_p {\rm Cov}(\boldsymbol{S})^{-1}$, whose elements are constant.

Since $g(S)$ is uncorrelated with $S$, $\epsilon$ is independent of $S$, we have $E(\boldsymbol{S}(g(\boldsymbol{S})+\epsilon))=E(\boldsymbol{S})E(g(\boldsymbol{S})+\epsilon)=\boldsymbol{0}$. Then according to central limit theorem, we have $\boldsymbol{W}_s\rightarrow_d N(\boldsymbol{0}, {\rm Cov}(\boldsymbol{S}(g(\boldsymbol{S})+\epsilon)))$, where $\rightarrow_d$ denotes converge in distribution.

For the covariance term, we have
\begin{equation}
\begin{aligned}
    {\rm Cov}(\boldsymbol{S}(g(\boldsymbol{S})+\epsilon)) 
    &= E((g(\boldsymbol{S})+\epsilon)^2 \boldsymbol{S}\boldsymbol{S}^T) \\ 
    &= E(g^2(\boldsymbol{S})\boldsymbol{S}\boldsymbol{S}^T)+\sigma^2 {\rm Cov}(\boldsymbol{S}) \\
    &\leq \delta^2 B^2 \boldsymbol{1}_{p_s\times p_s}+\sigma^2 {\rm Cov}(\boldsymbol{S}) \\ \nonumber
\end{aligned}
\end{equation}

According to the convergence property of random variables, now we have 
\begin{equation}
\begin{aligned}
\boldsymbol{\zeta}&\rightarrow_d {\rm Cov}(\boldsymbol{S})^{-1} N(\boldsymbol{0}, {\rm Cov}(\boldsymbol{S}(g(\boldsymbol{S})+\epsilon)))\\
&= N(\boldsymbol{0}, {\rm Cov}(\boldsymbol{S})^{-1}{\rm Cov}(\boldsymbol{S}(g(\boldsymbol{S})+\epsilon)){\rm Cov}(\boldsymbol{S})^{-1}) \\ \nonumber
\end{aligned}
\end{equation}

where the covariance term is bounded by $\delta^2 B^2{\rm Cov}(\boldsymbol{S})^{-1}\boldsymbol{1}_{p_s\times p_s}{\rm Cov}(\boldsymbol{S})^{-1}+\sigma^2{\rm Cov}(\boldsymbol{S})^{-1}$.
Therefore, $\zeta_i$ converges in distribution to a Gaussian random variable with finite variance.
Similarly, we can prove that $\xi_i$ also shares the same property, where the variance of the first term is bounded by  $\delta^2 B^2 {\rm Cov}(\boldsymbol{V}, \boldsymbol{S}) {\rm Cov}(\boldsymbol{S})^{-1}\boldsymbol{1}_{p_s\times p_s}{\rm Cov}(\boldsymbol{S})^{-1}{\rm Cov}(\boldsymbol{S}, \boldsymbol{V})+\sigma^2{\rm Cov}(\boldsymbol{V}, \boldsymbol{S}) {\rm Cov}(\boldsymbol{S})^{-1}{\rm Cov}(\boldsymbol{S}, \boldsymbol{V})$, and variance of the second term is bounded by $\sigma^2 B^2 \boldsymbol{1}_{p_s\times p_s}+\sigma^2 {\rm Cov}(\boldsymbol{V})$.  

We denote $M^2$ as the common upper bound of their variance, where $M=h(\delta, B, {\rm Cov}(\boldsymbol{X}))$ is some positive scalar which can be written as a function of $\delta, B, {\rm Cov}(\boldsymbol{X})$ unrelated to $n$. 

We will make use of the tail probability of Gaussian distribution that for $t>0$,
\begin{equation}
    1-\Phi(t)<t^{-1}e^{-\frac{1}{2}t^2}.
\end{equation}

Since $\lambda_n=o(n), c<1$, when $n\rightarrow \infty$
\begin{equation}
\begin{aligned}
& \quad \sum_{i=1}^{p_s}P(|\zeta_i^{(n)}|\geq \sqrt{n}(|\beta_i|-\frac{\lambda_n}{2n}b_i^{(n)})\\
&\leq \sum_{i=1}^{p_s}P(1-\Phi(\frac{\sqrt{n}(|\beta_i|-\frac{\lambda_n}{2n}b_i^{(n)}}{M})\\
&=\sum_{i=1}^{p_s}P(1-\Phi(\frac{\sqrt{n}(|\beta_i|+o(1))}{M})\\
&\leq \sum_{i=1}^{p_s} \frac{M}{n^{\frac{1}{2}}|\boldsymbol{\beta}_i|}e^{-\frac{|\boldsymbol{\beta_i}|^2 }{2M^2}n}\\
&=o(e^{-n^c})\\ \nonumber
\end{aligned}
\end{equation}

Since $\lambda_n=\omega(n^{\frac{1+c}{2}})$, we have $\frac{\lambda_n}{\sqrt{n}}=\omega(n^{\frac{c}{2}})$.

When $n\rightarrow \infty$
\begin{equation}
\begin{aligned}
& \quad \sum_{i=1}^{p_v}P(|\xi_i^{(n)}|\geq \frac{\lambda_n}{2\sqrt{n}}\eta_i) \\ 
&\leq \sum_{i=1}^{p_v}(1-\Phi(\frac{\lambda_n \eta_i}{2M\sqrt{n}}))\\
&\leq \sum_{i=1}^{p_v}\frac{2s\eta_i}{\frac{\lambda_n}{\sqrt{n}}}e^{-\frac{\eta_i^2}{8M^2}(\frac{\lambda_n}{\sqrt{n}})^2}\\
&=\sum_{i=1}^{p_v}o(n^{-\frac{c}{2}}e^{-\frac{\eta_i^2}{8M^2}n^c})\\
&=o(n^{-\frac{c}{2}}e^{-\frac{\min_{i=1}^{p_v}\{\eta_i\}^2}{8M^2}n^c})\\ \nonumber
\end{aligned}
\end{equation}

Since $M\geq \frac{\min_{i=1}^{p_v}\{\eta_i\}}{2\sqrt{2}}$, we have:
\begin{equation}
\frac{o(e^{-n^c})}{o(n^{-\frac{c}{2}}e^{-\frac{\min_{i=1}^{p_v}\{\eta_i\}^2}{8M^2}n^c})}\rightarrow 0 \nonumber    
\end{equation}
Thus we have:
\begin{equation}
    P(\hat{\boldsymbol{\beta}}(\lambda_n)=_s \boldsymbol{\beta})\geq 1-o(n^{-\frac{c}{2}}e^{-\frac{\min_{i=1}^{p_v}\{\eta_i\}^2}{8M^2}n^c}). \nonumber
\end{equation}

\subsection{Training Details}

We generate $\boldsymbol{X}=\{\boldsymbol{S}, \boldsymbol{V}\}$ from a multivariate Gaussian distribution $\boldsymbol{X}\sim N(\boldsymbol{0}, \boldsymbol{\Sigma})$.
In this way, we can simulate different correlation structures of $\boldsymbol{X}$ through the control of covariance matrix $\boldsymbol{\Sigma}$. 
In our experiments, we tend to make the correlations inside stable variables strong.
For simplicity, We further set $\boldsymbol{\Sigma}=Diag \left( \boldsymbol{\Sigma}^{(S)}, \boldsymbol{\Sigma}^{(V)} \right)$, assuming block structure. 
For $\boldsymbol{\Sigma}^{(S)} \in \mathbb{R}^{p_s \times p_s}$, its elements are: $\boldsymbol{\Sigma}^{(S)}_{jk}=\rho_s$ for $j \neq k$ and $\boldsymbol{\Sigma}^{(S)}_{jk}=1$ for $j=k$. 
For $\boldsymbol{\Sigma}^{(V)} \in \mathbb{R}^{p_v \times p_v}$, its elements are: $\boldsymbol{\Sigma}^{(V)}_{jk}=\rho_v$ for $j \neq k$ and $\boldsymbol{\Sigma}^{(V)}_{jk}=1$ for $j=k$. 
Despite we can simulate more complex settings by manipulating the covariance matrix $\boldsymbol{\Sigma}$, such a simplified setting is enough since it can adjust correlation inside $\boldsymbol{S}$ or inside $\boldsymbol{V}$. 
In our experiments, to simulate the scenario where correlations inside stable variables are strong, we tend to set $\rho_s$ a large value. We set $\rho_s=0.9$ and $\rho_v=0.1$ in our experiments.

For linear and nonlinear settings, we employ different data generation functions. It is worth noting that a certain degree of model misspecification is needed, otherwise the model will be able to directly learn the true stable variables using simply OLS or MLP. 

For the linear setting, we introduce the model misspecification error by adding an extra polynomial term to the dominated linear term, and later use a linear model to fit the data. The generation function is as follows:
\begin{equation}
\small
    Y=f(\boldsymbol{S})+\epsilon=[\boldsymbol{S}, \boldsymbol{V}]\cdot [\boldsymbol{\beta}_s, \boldsymbol{\beta}_v]^T + \boldsymbol{S}_{\cdot, 1}\boldsymbol{S}_{\cdot, 2}\boldsymbol{S}_{\cdot, 3} + \epsilon
\end{equation}
where $\boldsymbol{\beta}_s=\{\frac{1}{3}, -\frac{2}{3}, 1, -\frac{1}{3}, \frac{2}{3}, -1,... \}, \boldsymbol{\beta}_v=\boldsymbol{0}, \epsilon \sim \mathcal{N}(0, 0.3)$. 

For the nonlinear setting, we generate data in a totally nonlinear fashion. We employ random initialized MLP as the data generation function.
\begin{equation}
\small
    Y=f(\boldsymbol{S})+\epsilon = MLP(\boldsymbol{S})+\epsilon 
\end{equation}
Later we use MLP with a smaller capacity to fit the data. 
where the MLP is 2-layer with 100 hidden units, $\epsilon\sim \mathcal{N}(0, 0.1)$. To introduce model misspecification, we use a 2-layer MLP with 10 hidden units to fit the data after employing SVI to do variable selection. Parameters of MLP are initialized with uniform distribution $U(-1,1)$.

\begin{figure*}[t]
	\centering
	\subfigure[\#iterations = 0] {
	\label{fig:heat0}
	    \includegraphics[width=0.45\linewidth]{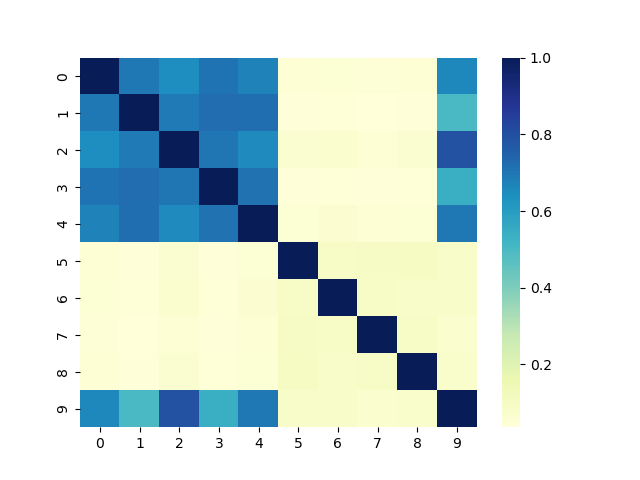}
	}
	\subfigure[\#iterations = 3] {
	\label{fig:heat3}
	    \includegraphics[width=0.45\linewidth]{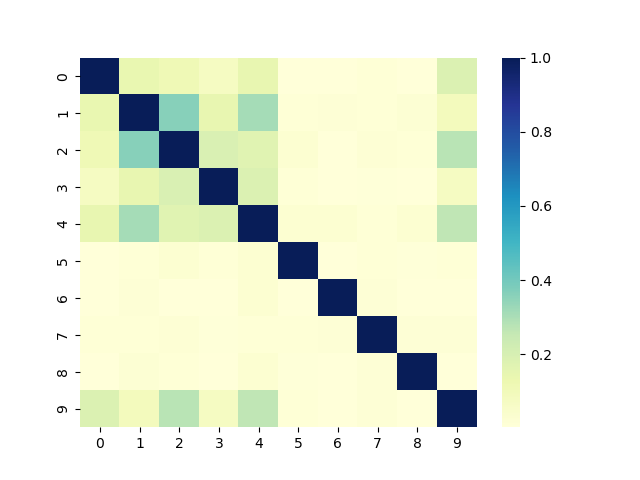}
	}
\end{figure*}

\begin{figure*}[!ht]
	\centering
	\subfigure[\#iterations = 7] {
	\label{fig:heat7}
	    \includegraphics[width=0.45\linewidth]{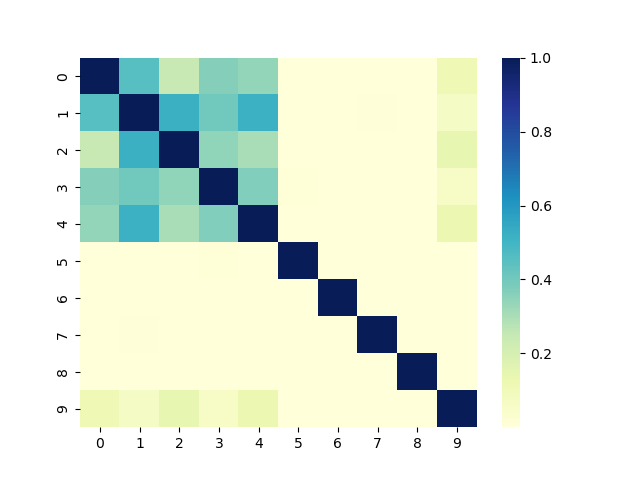}
	}
	\subfigure[\#iterations = 10] {
	\label{fig:heat10}
	    \includegraphics[width=0.45\linewidth]{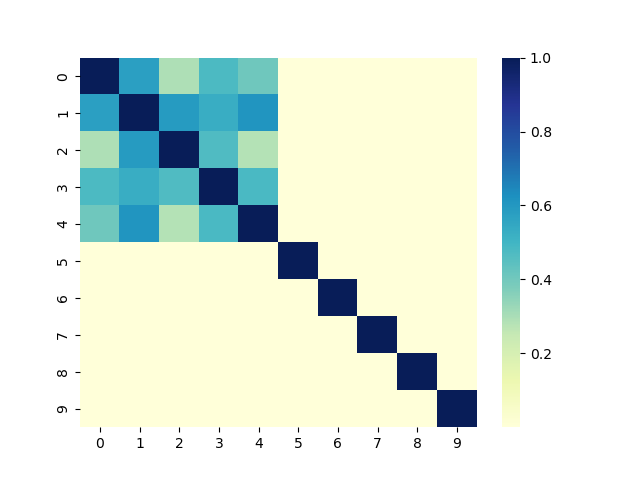}
	}
	\caption{Correlation heatmap with growing \#iterations, when fixing $r_{train}=2.5,n=3000$}
	\label{fig:heatmap}
\end{figure*}

\begin{figure*}[t]
	\centering
	\subfigure[Sample size $n=2000$] {
	\label{fig:period_n2000}
	    \includegraphics[width=0.32\linewidth]{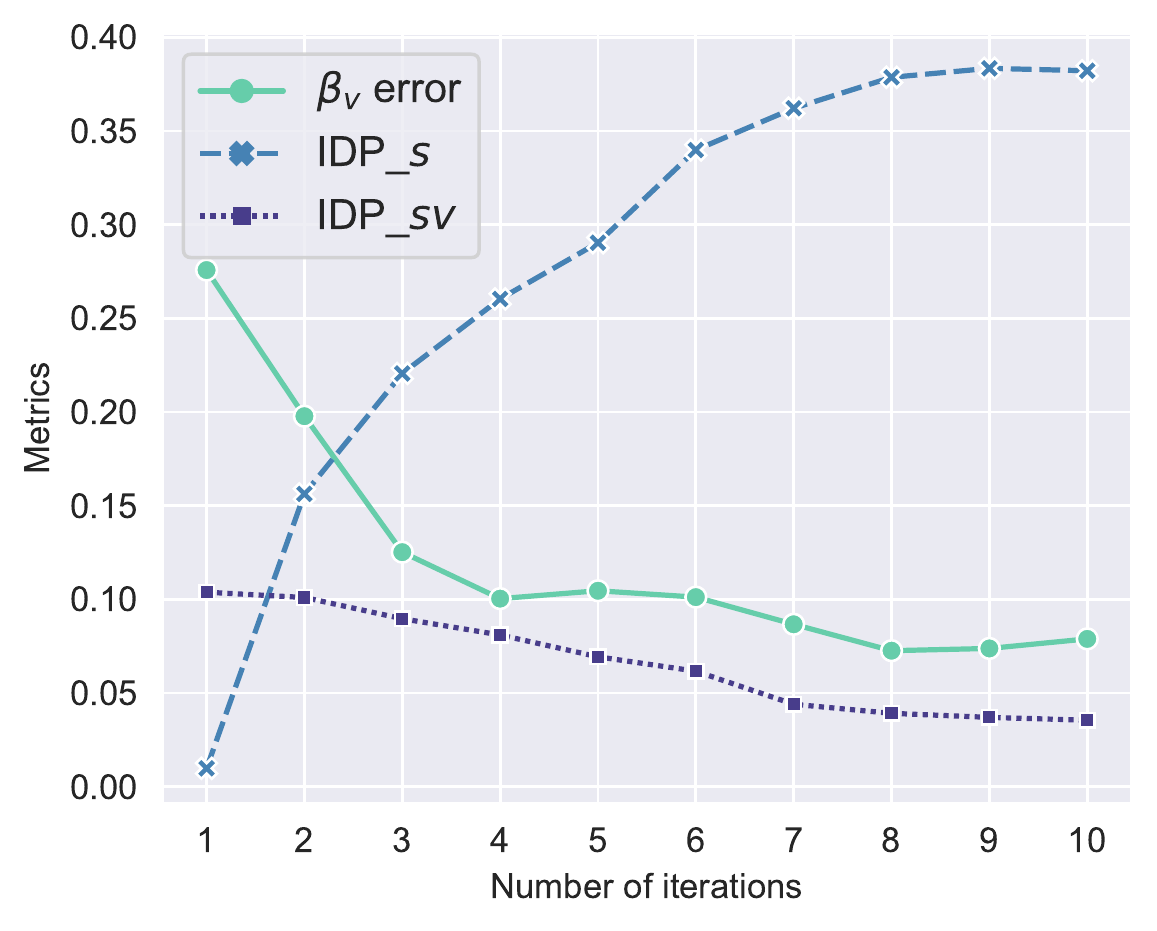}
	}
	\subfigure[Sample size $n=3000$] {
	\label{fig:period_n3000}
	    \includegraphics[width=0.32\linewidth]{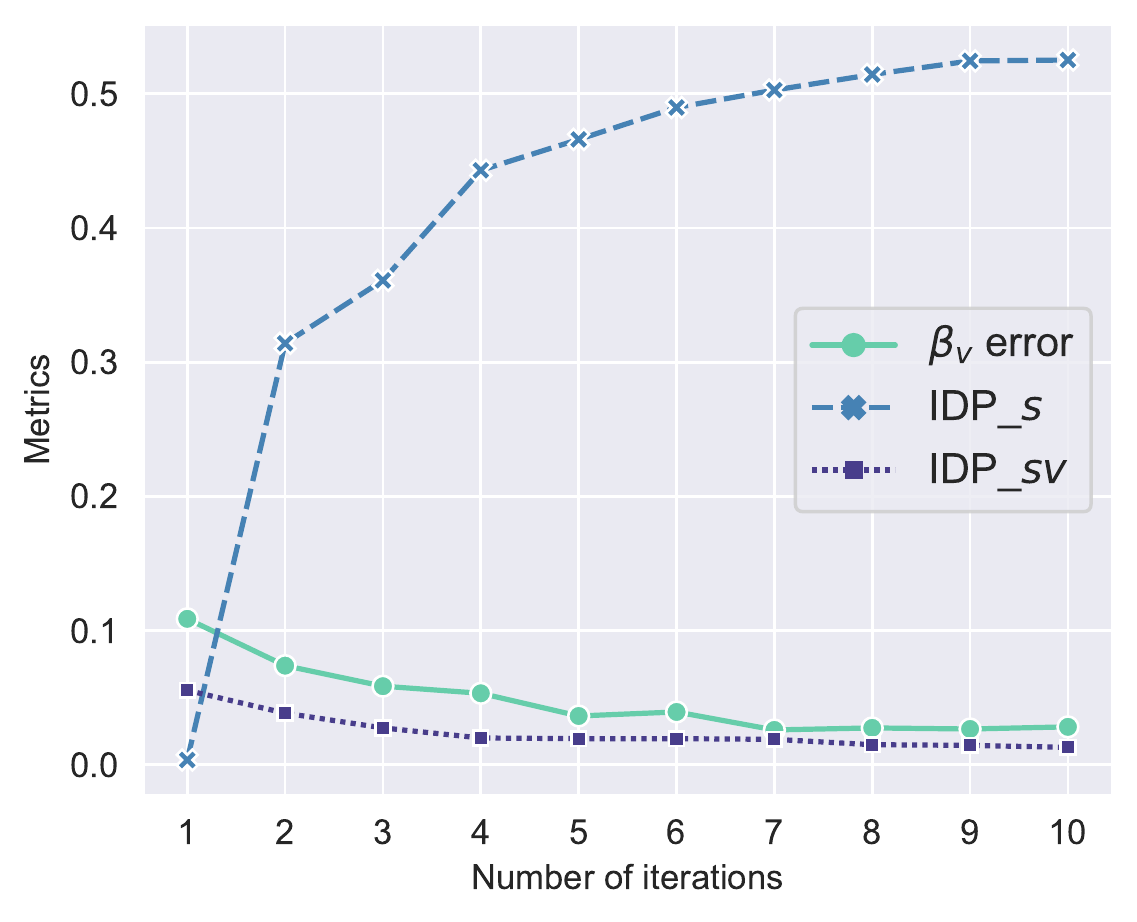}
	}
	\subfigure[Sample size $n=4000$] {
	\label{fig:period_n4000}
	    \includegraphics[width=0.32\linewidth]{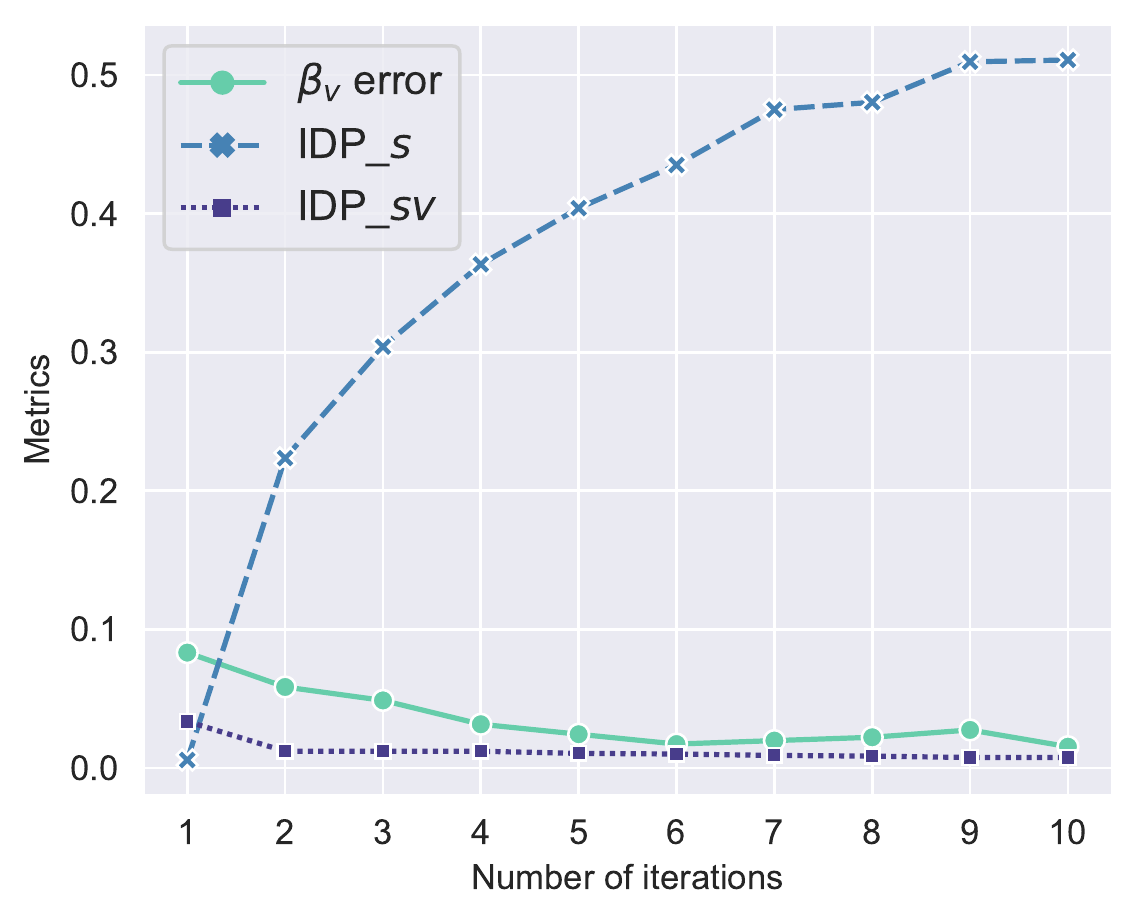}
	}
	\caption{$||\beta_v||_1, {\rm IDP}_s, {\rm IDP}_{sv}$ with growing \#iterations, when fixing $r_{train}=2.5.$}
	\label{fig:period_n}
\end{figure*}

\subsection{Iteration and convergence analysis}

As we have mentioned, it is rather difficult to analyze the optimization process of iterative algorithms and their convergence \cite{liu2021heterogeneous, zhou2022model} from a theoretical perspective. Therefore, we conduct additional experiments to illustrate empirically. 
\begin{itemize}
    \item Figure \ref{fig:heatmap} presents reweighted correlation heatmaps with the growing number of iterations. Among the 10 covariates, the first 5 dimensions are stable variables $\boldsymbol{S}$. The last 5 dimensions are unstable variables $\boldsymbol{V}$, and the last dimension is the biased unstable variable $V_b$. In Figure \ref{fig:heat0}, i.e. the original correlation heatmap without reweighting, there exist strong correlations inside $\boldsymbol{S}$, and the biased $V_b$ is also strongly correlated with $\boldsymbol{S}$, which exerts negative effects on covariate-shift generalization. In Figure \ref{fig:heat3} and \ref{fig:heat7}, we find that the correlations inside $\boldsymbol{S}$ (the top-left corner) are initially weakened a lot, but gradually they are strengthened. Besides, the correlations between $\boldsymbol{S}$ and $\boldsymbol{V}$ are weakened. In Figure \ref{fig:heat10}, at the end of the iterative process, a large proportion of the correlations inside $\boldsymbol{S}$ are preserved compared with Figure \ref{fig:heat3} and \ref{fig:heat7}. Meanwhile, $\boldsymbol{S}$ and $\boldsymbol{V}$ are decorrelated thoroughly. 
    This demonstrates that our proposed SVI succeeds in focusing on removing correlations between stable variables and unstable variables while avoiding decorrelating inside stable variables with the iterative process going on. 
    \item Figure \ref{fig:period_n} illustrates the iterative process and convergence of SVI with the help of three metrics: $\beta_v$ error, IDP$_s$, and IDP$_{sv}$. For IDP$_s$, we calculate the mean of off-diagonal elements in the reweighted correlation matrix of $\boldsymbol{S}$. For IDP$_{sv}$, we calculate the mean of elements in the reweighted correlation matrix between $\boldsymbol{S}$ and $\boldsymbol{V}$ \footnote{When drawing the lineplots, IDP$_{sv}$ is rescaled to match the magnitude of other two metrics for convenience}. We use these two metrics to roughly depict the independence between variables. 
    As we can see, the increase of IDP$_s$ implies that the correlations inside $\boldsymbol{S}$ are preserved better, and the decrease of IDP$_{sv}$ implies that the correlations between $\boldsymbol{S}$ and $\boldsymbol{V}$ are better removed. Both IDP$_s$ and IDP$_{sv}$ demonstrate better frontend sample reweighting with the iteration going on. Besides, the gradual decline of $\beta_v$ error shows the improvement of the backend sparse learning module brought by the iterative procedure. This proves that the iterative procedure truly helps frontend sample reweighting and backend sparse learning benefit each other. Meanwhile, all these three metrics show a trend of convergence, which could serve as empirical evidence of SVI's convergence.
\end{itemize}

\subsection{Real-world Dataset Description}

\subsubsection{House Price Prediction}

This task comes from a real-world dataset of houses from King County, USA. We acquire this dataset from Kaggle\footnote{\url{https://www.kaggle.com/c/house-prices-advanced-regression-techniques/data}}. 
It is a regression task, where the outcome is the house price and each sample contains 20 variables as potential inputs, such as the number of bedrooms or bathrooms. We first drop 3 useless columns like zip-code. Then we simulate different environments according to the built year, dividing the dataset into 6 periods. Each period is approximately two decades. To examine the covariate-shift generalization ability, we train on the first environment in which houses were built between 1900 and 1919, and test on the other 5 periods. 

\subsubsection{People Income Prediction}
In this experiment we employ the Adult dataset \cite{kohavi1996scaling}, formulating a binary classification task. The target variable is whether an adult's income per year exceeds 50K. We notice that the variables race and sex naturally produce sub-populations in the dataset, thus we use a combination of race and sex to generate 10 environments. 
We train on the first environment $(White, Female)$ and test on the other 9 environments to simulate distribution shift.

\subsection{Optimization and Analysis}
\label{sec:optim}

For the optimization of DWR loss, the learning process of MLP classifier in SRDO's density ratio estimation, and the optimization of sparse learning loss, we employ gradient descent. 
Besides, when conducting the last step in SVI, it is free to set the threshold $\gamma$ in a relatively wide range, since empirically we find that elements in the soft mask $\boldsymbol{\mu}$ are well separated after being optimized to convergence. 

With respect to the time complexity of our algorithm, for SVI of linear version, the complexity of frontend DWR is $O(np^2) $\cite{kuang2020stable}, where $n$ is sample size and $p$ is the dimension of input. The complexity of the backend sparse learning module is $O(np)$. Thus the total complexity is $O(Tnp^2)$, where $T$ is the number of iterations. 

For NonlinearSVI, in SRDO, the complexity of random permutation is $O(n)$, time cost of calculating the binary classification loss and updating it for density estimation is $O(np)$. The complexity of backend is also $O(np)$. Thus the total complexity is $O(np)$.

\subsection{Related Work}

In this section, we investigate several strands of works, including domain adaptation, domain generalization, invariant learning, and stable learning. 

% DA
\textbf{Domain adaptation} (DA) is the earliest stream of literature seeking to solve the problem of generalization under distribution shift. 
Some conduct importance weighting to match training and test distribution \cite{huang2006correcting, shimodaira2000improving, chu2016selective}. 
Some employ representation learning to transform raw data into a variable space where the discrepancy between source and target distribution is smaller \cite{fernando2013unsupervised, long2015learning, ganin2015unsupervised, tzeng2017adversarial} with theoretical support \cite{ben2010theory, ben2006analysis}.  
Nevertheless, on one hand, the availability of unlabeled test data is not quite realistic in that we can hardly know what kind of distribution shift will exist in real test environments. 
On the other hand, such methods only guarantee effectiveness under a certain test distribution, thus it may still suffer from performance degradation on other possible test distributions. 

% DG 
\textbf{Domain generalization} (DG) provides a solution for OOD generalization where we do not have access to test data apriori. 
It typically requires multiple subpopulations (namely domains) in training data and learns representations that are invariant across these subpopulations \cite{ghifary2015domain, li2018domain, dou2019domain}. 
Other techniques include meta learning \cite{li2018learning, balaji2018metareg, li2019episodic} and augmentation of source domain data \cite{carlucci2019domain, shankar2018generalizing, qiao2020learning}. 
However, theoretical support is absent in domain generalization, and empirically it strongly relies on the abundance of training domains. More recently, there are doubts that simple ERM performs comparably with popular DG methods if carefully tuned \cite{gulrajani2020search}. 

% invariant learning
\textbf{Invariant learning} is another brunch of works, which develop theories to support the proposed algorithms. 
Similar to domain generalization, it tries to capture the invariant part across different training environments to increase models' generalization ability on unknown test distribution, thus it still requires either explicit environment labels \cite{arjovsky2019invariant, koyama2020out, krueger2021out} or assumes enough heterogeneity in training data \cite{liu2021heterogeneous, liu2021kernelized}. 

% stable learning
\textbf{Stable learning} focuses on covariate shift with a structural assumption on covariates. It usually does not require explicit environmental information \cite{cui2022stable}.
\cite{shen2018causally, kuang2018stable} apply global balancing of moments under the setting where input variables and the outcome variable are binary.
\cite{shen2020stable, kuang2020stable} focus on the linear regression task with model misspecification by global decorrelation after sample reweighting. 
Recently, \cite{zhang2021deep} extends it to deep learning empirically. 
In \cite{xu2021stable}, they conduct a theoretical analysis of stable learning from the perspective of variable selection, proving that perfect reweighting and variable decorrelation help select stable variables regardless of whether the data generation mechanism is linear or nonlinear. However, under the finite-sample setting, this can hardly be achieved.

% \clearpage
\bibliography{aaai23}

\end{document}